\documentclass[lettersize,journal]{IEEEtran}
\usepackage{amsmath,amsfonts}
\usepackage{algorithmic}
\usepackage{algorithm}
\usepackage{array}
\usepackage[caption=false,font=normalsize,labelfont=sf,textfont=sf]{subfig}
\usepackage{textcomp}
\usepackage{stfloats}
\usepackage{url}
\usepackage{verbatim}
\usepackage{graphicx}
\usepackage{cite}
\usepackage[table]{xcolor}
\usepackage{booktabs}
\usepackage{multirow}
\usepackage{makecell}
\usepackage{threeparttable}
\usepackage{graphicx}
\usepackage[margin=1in]{geometry}
\usepackage{float}
\usepackage[hidelinks]{hyperref}
\usepackage{xcolor}   
\usepackage{soul}   
\definecolor{best}{RGB}{248,204,204}      
\definecolor{second}{RGB}{252,229,205}    
\definecolor{third}{RGB}{255,250,204}     
\sethlcolor{best}
\sethlcolor{second}
\sethlcolor{third}

\usepackage{caption}
\usepackage{graphicx}
\usepackage{xcolor}     
\usepackage{tikz}       
\usepackage[absolute,overlay]{textpos} 

\begin{document}

\title{ConFi-GS: Confidence-Guided High-Frequency Injection for 3D Gaussian Splatting Super-Resolution}

\author{Jiaxiang Li, Zongtan Zhou, Zhen Tan, Yadong Liu, Dewen Hu,~\IEEEmembership{Senior Member,~IEEE,}
}

\markboth{Journal of \LaTeX\ Class Files,~Vol.xx~x, No.~x, xxxx~202x}%
{Shell \MakeLowercase{\textit{et al.}}: A Sample Article Using IEEEtran.cls for IEEE Journals}

\IEEEpubid{}

\maketitle

\begin{abstract}
Reconstructing high-quality 3D scenes from low-resolution multi-view images remains challenging for 3D Gaussian Splatting (3DGS), because insufficient high-frequency observations often lead to blurred textures, weak boundaries, and view-inconsistent details. Existing approaches either apply super-resolution guidance uniformly or localize enhancement regions based mainly on geometric sampling. However, they typically do not distinguish between two fundamentally different questions: where additional detail is needed, and whether the corresponding candidate high-frequency content is reliable enough to be internalized into a multi-view consistent 3D representation. 

In this paper, we propose a reliability-aware frequency modeling framework for low-resolution 3DGS reconstruction. The framework first estimates a geometry-guided detail-demand prior to locate regions that are likely under-detailed under low-resolution supervision. It then computes a frequency-aware reliability map to determine whether candidate high-frequency details are structurally supported, spectrally unresolved, and cross-view stable. Combining these signals yields a detail-injection map that guides where super-resolved details should be introduced during optimization. Based on this map, we design a unified optimization scheme comprising spatially selective supervision, coarse-to-fine frequency regularization, and reliability-aware Gaussian densification. This scheme controls where reliable details are injected, when high-frequency supervision is activated, and how unresolved yet reliable details are internalized into the Gaussian representation. Experiments on multiple benchmarks show improved fidelity and perceptual quality while suppressing unstable or view-inconsistent details.

\end{abstract}

\begin{IEEEkeywords}
3D Gaussian Splatting, low-resolution multi-view reconstruction, 3D super-resolution, reliability-aware detail injection, frequency-aware regularization
\end{IEEEkeywords}

\section{Introduction}

Recent advances in novel view synthesis, digital twins, immersive interaction, and high-quality 3D content creation have increased the demand for efficient and accurate scene representations. Traditional implicit neural radiance field (NeRF) methods~\cite{mildenhall2020nerf,muller2022instant,FoVNeRF} can produce photorealistic renderings, but they often suffer from high computational cost and limited real-time performance. In contrast, 3D Gaussian Splatting (3DGS), as an explicit representation based on differentiable 3D Gaussian primitives and rasterization-based rendering, offers a favorable balance between rendering quality and efficiency~\cite{ref1,refsurvey,refsurvey2}, which has been further enhanced by spatially varying color primitives~\cite{SVGS}. This has made 3DGS an attractive choice for multi-view reconstruction and novel view synthesis~\cite{refsurvey,refsurvey2,ref3,ref8}, while compact multi-plane representations have been explored to improve storage efficiency~\cite{MPGS}, and foveated strategies have extended it to immersive dynamic scene rendering~\cite{FovGS}.

Despite these advantages, the quality of 3DGS reconstruction remains highly dependent on the quality of the input images. In many practical scenarios, multi-view observations are captured at limited spatial resolution due to sensor, storage, or transmission constraints. When only low-resolution images are available, the learned Gaussian representation is supervised mainly by coarse visual evidence, which often leads to blurred textures, weak boundaries, and unstable fine structures in high-resolution renderings. This limitation has become a major obstacle to applying 3DGS in scenarios that require visually sharp and structurally stable 3D reconstruction.

A common strategy for addressing this issue is to introduce external super-resolution priors. However, low-resolution 3DGS reconstruction involves a fundamental ambiguity: a region that appears under-detailed does not necessarily provide reliable high-frequency evidence for 3D optimization. In other words, detail demand and detail reliability are not the same. A region may require additional detail under low-resolution supervision, yet the candidate high-frequency content provided by a super-resolved reference can still be view-specific, structurally unsupported, or inconsistent across nearby views. Directly enforcing such details may improve local sharpness in individual images while simultaneously harming multi-view consistency and structural stability of the learned 3D representation.

Existing methods address this problem only partially. Anti-aliasing and scale-aware rendering methods improve robustness across resolutions, but they mainly reconstruct observed information more faithfully rather than recover missing high-frequency content. Methods that incorporate super-resolution priors can introduce richer candidate details, yet they often treat those details as broadly useful supervision and remain vulnerable to cross-view inconsistency~\cite{FISN}. Geometry-guided selection strategies help localize regions where enhancement is more likely to be needed, but they primarily answer where detail may be missing rather than whether the corresponding high-frequency evidence can be trusted~\cite{GeoTexDensifier}. Frequency-based regularization provides a principled mechanism for progressive spectral refinement, but without an explicit reliability assessment, unstable pseudo-high-frequency signals may still be absorbed into the 3D representation. In addition, recent optimization and structure-aware approaches attempt to improve reconstruction robustness and detail preservation, such as learned optimization strategies for sparse views~\cite{LeOpGS} and semantic-guided enhancement of important structures~\cite{SGGS}, yet they do not explicitly address the reliability of high-frequency supervision under low-resolution conditions. 

In this paper, we formulate low-resolution 3DGS reconstruction as a reliability-aware frequency modeling problem. Our key observation is that the reconstruction process should explicitly separate two questions: where additional detail is needed, and whether the candidate high-frequency content in those regions is reliable enough to be internalized into a multi-view consistent 3D scene representation. To this end, we first estimate a geometry-guided detail-demand prior that identifies under-detailed regions from a cross-view sampling perspective. We then compute a frequency-aware reliability map that evaluates whether the candidate high-frequency content is structurally supported, spectrally unresolved, and cross-view stable. By combining these two signals, we obtain a final detail-injection map, which governs where super-resolved detail should be introduced into optimization.

Based on this formulation, we further develop a unified optimization framework in which the detail-injection map guides spatially selective supervision, coarse-to-fine frequency regularization, and reliability-aware Gaussian densification. Rather than treating super-resolved details as uniformly beneficial, our method introduces them only when they are both needed and sufficiently reliable, and progressively internalizes unresolved reliable detail into the Gaussian representation through structural adaptation. In this way, the proposed method shifts low-resolution 3DGS enhancement from direct detail amplification toward reliability-aware detail internalization.

The main contributions of this paper are summarized as follows:

1. Reliability-aware problem formulation for low-resolution 3DGS.We reformulate low-resolution multi-view 3D Gaussian Splatting as a reliability-aware detail-injection problem, explicitly separating detail demand from detail reliability.

2. Geometry-guided demand estimation and frequency-aware reliability assessment.
We first estimate a geometry-guided detail-demand prior and then refine it with a frequency-aware reliability assessment based on structural support, unresolved high-frequency evidence, and cross-view consistency, producing a final detail-injection map for reliable supervision.

3. Unified optimization for reliable detail internalization.
We use the detail-injection map to jointly guide selective progressive frequency regularization and reliability-aware Gaussian densification, enabling reliable high-frequency detail to be introduced gradually and internalized structurally into the evolving 3D representation.

The rest of the paper is organized as follows. Section~\ref{sec:Related Work} reviews related work. Section~\ref{sec:Proposed method} presents the proposed framework. Section~\ref{sec:Experiments} reports experimental settings and results. Section~\ref{sec:Ablation Study} provides ablation studies and discussion. Section~\ref{sec:CONCLUSION} concludes the paper.

\section{Related Work}
\label{sec:Related Work}

Low-resolution 3D Gaussian Splatting (3DGS) is closely related to four research directions: scale-aware rendering, super-resolution-guided 3D reconstruction, geometry-guided selective enhancement, and frequency-aware modeling. Existing methods improve reconstruction quality under degraded observations from different perspectives, but they typically focus on either providing richer candidate details, localizing where enhancement is needed, or regulating how frequency information is introduced during optimization. In contrast, our work focuses on a complementary question: whether the candidate high-frequency content is reliable enough to be internalized into a multi-view consistent 3D representation.

\subsection{Scale-Aware Rendering for 3DGS}

A first line of work improves the robustness of 3DGS across rendering scales through anti-aliasing and scale-aware modeling. The original 3DGS framework demonstrates the efficiency and rendering quality of explicit Gaussian representations~\cite{ref1}, while subsequent works extend it to view-adaptive rendering~\cite{lu2024scaffold}, reflective surfaces~\cite{jiang2024gaussianshader}, and dynamic scenes~\cite{yang2024deformable,FUGS}. Early anti-aliased neural representations such as Mip-NeRF introduced cone-based sampling to reduce aliasing before per-view ray casting~\cite{barron2021mipnerf}, which was later extended to unbounded scenes in Mip-NeRF 360~\cite{barron2022mipnerf360}. Complementary to NeRF-based anti-aliasing, 2D Gaussian Splatting achieves more accurate geometric reconstruction through oriented 2D disks instead of isotropic 3D ellipsoids~\cite{huang20242dgs}, while Mip-Splatting adapts similar multiscale ideas to the Gaussian representation to suppress aliasing across different rendering resolutions~\cite{yu2024mipsplatting}. These methods improve rendering stability and better preserve observed scene content across scales. However, they primarily aim to reconstruct the available observations more faithfully, rather than recover missing high-frequency detail from low-resolution inputs. As a result, they improve scale robustness but do not directly address the problem of reliable detail recovery under low-resolution supervision.

\subsection{Super-Resolution Priors for Low-Resolution 3D Reconstruction}

Another important direction introduces external super-resolution (SR) priors into 3D reconstruction. Recent advances in single-image super-resolution, including CNN-based methods~\cite{dong2014srcnn,lim2017edsr,zhang2018rcan}, GAN-based approaches~\cite{ledig2017srgan,wang2021realesrgan}, transformer architectures~\cite{liang2021swinir}, and blind SR methods that model realistic degradation chains for complex real-world inputs~\cite{zhang2021bsrgan}, provide strong 2D priors for detail recovery. Methods such as SRGS, GaussianSR, SuperGaussian, S2Gaussian, and Sequence Matters show that these priors can substantially improve 3DGS reconstruction from low-resolution views~\cite{ref3,yu2024gaussiansr,shen2024supergaussian,ref8,ko2025sequencematters}. These methods demonstrate that external image priors can provide richer candidate high-frequency content than low-resolution observations alone.

Nevertheless, the use of SR priors also introduces a fundamental risk: visually plausible high-frequency details in individual views are not always trustworthy from a multi-view 3D perspective. Hallucinated textures, view-local artifacts, and structurally unsupported sharpness may improve per-view appearance while degrading cross-view consistency after being absorbed into the 3D representation. Most existing SR-guided 3DGS methods do not explicitly distinguish between regions that demand additional detail and regions whose candidate high-frequency content is sufficiently reliable for 3D optimization. Our work is motivated by this gap.

\subsection{Geometry-Guided Selective Enhancement}

Recent work has pointed out that SR guidance should not be applied uniformly across the image. Because the same 3D region can be projected at different scales in different views, some regions may still be relatively well sampled in a subset of observations, whereas others remain under-detailed throughout the dataset. SplatSuRe formalizes this idea by using scene geometry and camera poses to estimate under-sampled regions and selectively applying SR guidance where additional detail is more likely to be needed~\cite{asthana2025splatsure}.

This is an important step beyond uniform enhancement because it shifts the focus from whether to enhance to where enhancement is likely to be beneficial. However, geometry-guided selection mainly estimates detail demand. It does not fully determine whether the candidate high-frequency evidence in those regions can be safely trusted for multi-view consistent optimization. In difficult cases such as repeated textures, view-dependent appearance, or unstable SR predictions, regions with strong demand for enhancement do not necessarily coincide with regions whose candidate details are reliable. Our method builds on this line of work but explicitly separates geometry-guided detail-demand estimation from reliability-aware detail assessment.

\subsection{Frequency-Aware Modeling for 3DGS}

A closely related direction studies 3DGS from the frequency-domain perspective. FreGS shows that standard densification may overfit high-variance regions and addresses this issue using progressive frequency regularization and coarse-to-fine frequency annealing~\cite{zhang2024fregs}. More recent frequency-aware decomposition methods further suggest that spectral structure can help organize Gaussian representations and support more controllable detail refinement~\cite{lavi2025freqdecomp}. Beyond static scenes, frequency-aware uncertainty modeling has also been applied to dynamic scene reconstruction by explicitly modeling motion uncertainty in Fourier space~\cite{FUGS}.

These studies demonstrate that frequency information is valuable not only for measuring reconstruction discrepancy but also for regulating optimization and structural adaptation. Compared with purely geometry-based selection, frequency-aware approaches provide a more direct mechanism for controlling how high-frequency information enters the reconstruction process. However, most existing frequency-based methods do not explicitly address whether the candidate high-frequency signals themselves are reliable under low-resolution multi-view supervision. Without such a reliability assessment, unstable pseudo-high-frequency content may still be regularized, preserved, or structurally internalized even when it lacks sufficient support across views.

\subsection{Summary}

Overall, existing methods provide important but incomplete components for low-resolution 3DGS reconstruction. Scale-aware rendering improves robustness across resolutions, SR priors provide richer candidate detail, geometry-guided methods estimate where additional detail may be needed, and frequency-aware methods regulate how spectral information is introduced and refined. Our work is motivated by the gap between detail demand and detail reliability. Rather than treating geometric selection and frequency regularization as separate design choices, we connect them through a unified reliability-aware frequency modeling framework, in which geometry-guided demand estimation, frequency-aware reliability assessment, and selective detail internalization are jointly optimized for low-resolution 3DGS.

\section{Proposed Method}
\label{sec:Proposed method}

\subsection{Overview}

Given a low-resolution multi-view training set
\begin{equation}
\mathcal{D}=\{(I_t^{LR},\Pi_t)\}_{t=1}^{T}
\end{equation}
where $I_t^{LR}$ denotes the low-resolution observation at view $t$ and $\Pi_t$ is the corresponding camera parameter, we use a frozen single-image super-resolution model to generate a high-resolution reference for each training view:
\begin{equation}
\{I_t^{SR}\}_{t=1}^{T}
\end{equation}

Our goal is to optimize a high-resolution 3D Gaussian Splatting (3DGS) representation that preserves multi-view consistency while selectively internalizing only those high-frequency details that are both needed and reliable. The key challenge is that low-resolution supervision reveals where details are missing, but does not indicate whether the candidate high-frequency content suggested by $I_t^{SR}$ can be safely absorbed into a consistent 3D representation. In particular, super-resolved references may contain view-local artifacts, unstable pseudo-textures, or structurally unsupported sharp patterns. Therefore, detail demand and detail reliability should be modeled separately.

Based on this observation, we formulate low-resolution 3DGS reconstruction as a reliability-aware frequency modeling problem. Our method contains four coordinated components. First, we estimate a detail-demand map that identifies regions likely to lack recoverable detail under low-resolution supervision. Second, we compute a reliability map that evaluates whether the candidate high-frequency content in those regions is structurally supported, spectrally unresolved, and cross-view consistent. Third, we combine both signals into a final detail-injection map, which controls where reliable detail supervision should be applied. Fourth, we use this map to guide both selective progressive frequency regularization and reliability-aware Gaussian densification, so that reliable high-frequency information is introduced gradually and internalized structurally into the 3D representation.

\begin{figure*}[!t]
\centering
\includegraphics[width=\textwidth]{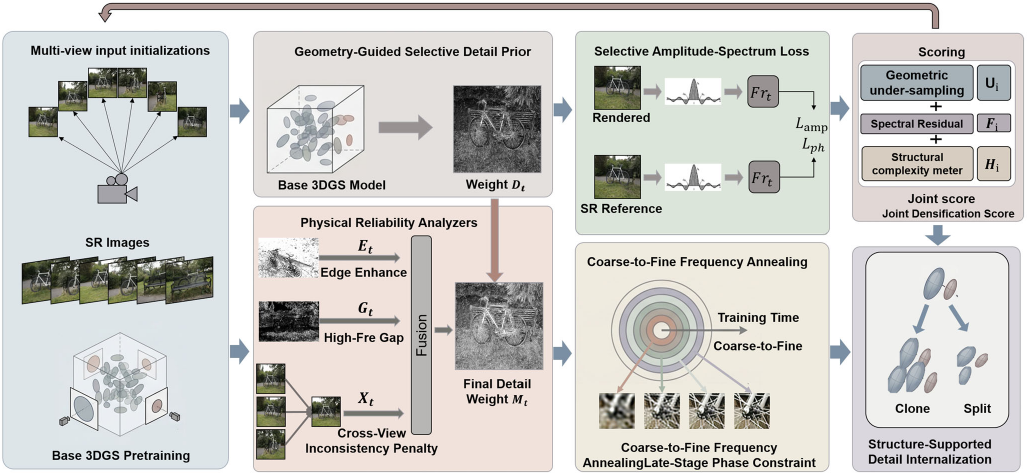}
\caption{Overview of the proposed reliability-aware frequency modeling framework for low-resolution 3DGS. Given low-resolution inputs and SR references, the framework estimates a detail-demand map, evaluates frequency-aware reliability, and produces a detail-injection map that guides selective frequency regularization and reliability-aware densification.}
\label{fig_main}
\end{figure*}

This design establishes a unified framework that controls where detail is needed, which candidate details can be trusted, when high-frequency supervision should be introduced, and how unresolved reliable details should trigger structural adaptation. The overall pipeline is illustrated in Fig.~\ref{fig_main}.

\subsection{Geometry-guided Detail Demand Estimation}

This module estimates where additional detail is needed from a geometric sampling perspective. It does not determine whether the candidate super-resolved details are trustworthy. We deliberately separate these two questions: geometric under-sampling indicates insufficient observation support, but it does not by itself imply that the corresponding high-frequency content in $I_t^{SR}$ is reliable.

We first pretrain a base 3DGS model using the low-resolution observations. For each Gaussian $G_i$ and training view $t$, we compute its screen-space projection radius as
\begin{equation}
r_i^t = 3\sqrt{\max(\lambda_1^i(t),\lambda_2^i(t))}
\end{equation}
where $\lambda_1^i(t)$ and $\lambda_2^i(t)$ are the two eigenvalues of the projected covariance matrix of $G_i$ in view $t$. We then collect the maximum and minimum projection radii over all visible views and define the cross-view sampling ratio
\begin{equation}
\rho_i=\frac{r_{i,\max}}{r_{i,\min}}
\end{equation}

This quantity measures how unevenly the same Gaussian is sampled across views. When $\rho_i$ is large, the Gaussian is observed at relatively higher fidelity in at least some views, so those views may provide transferable detail evidence for other observations. In contrast, when $\rho_i$ is close to $1$, the Gaussian is sampled at a similar, and often insufficient, scale across views, indicating a stronger need for external detail compensation.

To convert this geometric cue into a detail-demand score, we define
\begin{equation}
d_i = 1-\sigma\!\left(\frac{\rho_i-\tau}{k}\right)
\label{eq:detail-demand}
\end{equation}
where $\sigma(\cdot)$ denotes the Sigmoid function, $\tau$ controls the transition threshold, and $k$ controls the smoothness of the transition. Under this definition, a larger $d_i$ indicates a stronger demand for additional high-frequency supervision. This choice removes the ambiguity between sampling sufficiency and detail demand: highly under-sampled Gaussians receive larger demand scores, while Gaussians already well supported in some views receive smaller scores.

We rasterize the Gaussian-wise scores $\{d_i\}$ to each training view and obtain a view-specific detail-demand map
\begin{equation}
D_t \in [0,1]^{H\times W}
\end{equation}
$D_t$ provides a geometry-guided prior that highlights regions likely to be under-detailed. Importantly, $D_t$ should not be interpreted as a final supervision weight. A region may strongly demand additional detail while still lacking reliable high-frequency evidence. We therefore introduce a second stage to assess the reliability of the candidate details before applying them to optimization.

\subsection{Frequency-aware Reliability Assessment}

After identifying where detail is likely to be missing, we estimate whether the corresponding candidate high-frequency content is reliable enough to guide 3D optimization. The reliability assessment is designed to reject three common failure modes of pseudo-high-frequency supervision: structurally unsupported sharpness, unresolved but unstable spectral discrepancy, and view-specific details that cannot be stably reprojected across nearby views.

To this end, we compute a view-wise reliability map $C_t^{rel}$ from three complementary cues:
\begin{equation}
C^{rel}_t(p)=
\mathcal{N}
\left(
\sqrt{E_t(p)G_t(p)}\cdot (1-X_t(p))
\right)
\label{eq:reliability-map}
\end{equation}
where $\operatorname{\mathcal{N}}(\cdot)$ denotes image-wise normalization to $[0,1]$, $E_t$ measures structural support from local edges, $G_t$ measures unresolved high-frequency evidence, and $X_t$ penalizes cross-view inconsistency. We use different symbols for reliability and inconsistency to avoid semantic ambiguity.

The first cue $E_t$ is computed from the super-resolved reference image $I_t^{SR}$ and indicates whether candidate detail is aligned with locally salient structure. We convert $I_t^{SR}$ to grayscale and apply a Sobel operator:
\begin{equation}
E_t=\operatorname{Norm}\!\left(\left|\nabla I_t^{SR}\right|\right)
\end{equation}
\begin{equation}
\left|\nabla I_t^{SR}\right|=\sqrt{g_x^2+g_y^2+\varepsilon}
\end{equation}
where $g_x$ and $g_y$ are the horizontal and vertical gradient responses and $\varepsilon$ is a small constant for numerical stability. This term does not assume that all strong edges are reliable; rather, it provides structural support for candidate details that align with meaningful local boundaries or salient shape transitions.

The second cue $G_t$ measures how much high-frequency information in the reference is still missing from the current rendering:
\begin{equation}
G_t=\operatorname{Norm}\!\left(\left|\mathcal{H}(R_t^{HR})-\mathcal{H}(I_t^{SR})\right|\right)
\end{equation}
where $R_t^{HR}$ is the current high-resolution rendering and $\mathcal{H}(\cdot)$ denotes a Fourier-domain high-pass operator. In practice, $\mathcal{H}(\cdot)$ is implemented by applying a 2D Fourier transform, shifting the spectrum to the center, masking low frequencies with a radial high-pass filter, and transforming the response back to the spatial domain. The resulting magnitude map is averaged across channels to form a single-channel response. A larger $G_t$ indicates that the current 3DGS representation has not yet absorbed candidate high-frequency content suggested by the reference image. This cue alone does not imply reliability, but it identifies unresolved spectral evidence that may be worth injecting if supported by the other cues.

The third cue $X_t$ suppresses candidate details that are unstable across nearby views. For each target view $t$, we select a small neighboring view set $\mathcal{N}(t)$ according to camera proximity and reproject their high-frequency responses into the target view:
\begin{equation}
\hat{H}_{t}^{(k)}=\mathcal{W}_{k\rightarrow t}\!\left(\mathcal{H}(I_k^{SR})\right),  k\in\mathcal{N}(t)
\end{equation}
where $\mathcal{W}_{k\rightarrow t}$ denotes geometric reprojection from view $k$ to view $t$. We then measure the discrepancy between the target-view high-frequency response and the reprojected neighboring responses:
\begin{equation}
\Delta_t=
\frac{1}{|\mathcal{N}(t)|}
\sum_{k\in\mathcal{N}(t)}
\left|\mathcal{H}(I_t^{SR})-\hat{H}_{t}^{(k)}\right|
\end{equation}
Finally, we obtain the normalized inconsistency penalty
\begin{equation}
X_t=
\operatorname{clip}
\left(
\frac{\Delta_t}{Q_{0.95}(\Delta_t)+\varepsilon},
\,0,\,1
\right)
\end{equation}
where $Q_{0.95}(\Delta_t)$ is the $95\%$ quantile for robust normalization. A larger $X_t$ indicates that the candidate high-frequency pattern in $I_t^{SR}$ is less stable under reprojection and is therefore less trustworthy for 3D supervision.

We combine geometric demand and frequency-aware reliability into the final detail-injection map:
\begin{equation}
M_t=\operatorname{Norm}(D_t\odot C_t^{rel})
\label{eq:final-injection-map}
\end{equation}
where $\odot$ denotes element-wise multiplication. Under this definition, a region receives a high injection weight only when it satisfies both conditions: it demands additional detail and the candidate high-frequency evidence is sufficiently reliable. This formulation removes the internal inconsistency of using a sampling-sufficiency score directly as a supervision weight.

With Eq.~(\ref{eq:final-injection-map}), the final weight map no longer represents only geometric under-sampling. Instead, it explicitly selects regions that are simultaneously under-detailed, structurally supported, spectrally unresolved, and cross-view stable. In this way, the method reframes detail injection from a purely spatial selection problem into a reliability-aware filtering problem.

\subsection{Selective Progressive Frequency Regularization}

After obtaining the detail-injection map $M_t$, we introduce frequency supervision in a selective and progressive manner. The key idea is not to enforce full-band frequency matching from the beginning of training. Instead, we first stabilize coarse appearance and geometry, and only then gradually introduce higher-frequency constraints in regions where detail is both needed and reliable.

For each training view, we sample local patches from regions with high responses in $M_t$. Let $R_k$ and $I_k^{SR}$ denote the rendered patch and the corresponding super-resolved reference patch, respectively. Using a smooth normalized patch mask $W_k$ derived from $M_t$, we obtain
\begin{equation}
\hat R_k=R_k\odot W_k,\qquad
\hat I_k^{SR}=I_k^{SR}\odot W_k
\end{equation}
where $\odot$ denotes element-wise multiplication.

We then compute patch-level Fourier representations and define the amplitude-spectrum loss
\begin{equation}
L_{amp}=\frac{1}{K}\sum_{k=1}^{K}L_{amp}^{(k)}
\end{equation}
where $K$ is the number of selected patches. $L_{\mathrm{amp}}^{(k)}$ aligns the amplitude spectra of the masked rendered and SR patches under a coarse-to-fine frequency schedule. During training, the amplitude loss is activated progressively from low to high frequencies. This coarse-to-fine schedule avoids forcing the model to fit unstable pseudo-high-frequency details before the underlying geometry becomes sufficiently stable.

In addition, we introduce a phase-spectrum loss only at a later training stage:
\begin{equation}
L_{ph}=\frac{1}{K}\sum_{k=1}^{K}L_{ph}^{(k)}
\end{equation}
$L_{\mathrm{ph}}^{(k)}$ aligns their wrapped phase difference only within a valid frequency band with shared amplitude support. Compared with amplitude alignment, phase alignment is more sensitive to structural instability and local misregistration. We therefore activate $L_{ph}$ after the representation has become more stable, and only on valid frequency bands with shared spectral support.

Under this design, frequency regularization serves as a controlled detail-injection mechanism rather than a rigid full-image spectral constraint. The model first learns stable low-frequency structure, and then gradually absorbs reliable higher-frequency information according to $M_t$.

\subsection{Reliability-aware Gaussian Densification}

Loss-based supervision alone is insufficient to fully internalize reliable high-frequency details, because unresolved detail may remain limited by an overly coarse Gaussian layout. We therefore introduce a reliability-aware Gaussian densification strategy that allocates additional representational capacity according to detail demand, reliable spectral residual, and structural complexity.

For each Gaussian $G_i$, we define a joint densification score
\begin{equation}
S_i =
\left(
\bar{U}_i \bar{F}_i \bar{H}_i
\right)^{1/3}
\end{equation}
where $U_i$ measures geometric detail demand, $F_i$ measures the remaining reliable high-frequency residual associated with the Gaussian, and $H_i$ measures structural complexity through screen-space gradient statistics.

The term $U_i$ is inherited from the geometry-guided demand analysis and is directly set to
\begin{equation}
U_i=d_i
\end{equation}
Thus, Gaussians that remain under-supported across views receive higher densification priority.

The term $F_i$ aggregates the reliability-weighted high-frequency residual over the visible support region $\Omega_i^t$ of Gaussian $G_i$: 
\begin{equation} 
\Gamma_t(p)=\operatorname{Norm}\!\left( \left| \mathcal H(R_t^{HR})(p)-\mathcal H(I_t^{SR})(p) \right| \right) 
\end{equation} 
\begin{equation} 
F_i= \frac{\sum\limits_t\sum\limits_{p\in\Omega_i^t}M_t(p)\,\Gamma_t(p)} {\sum\limits_t\sum\limits_{p\in\Omega_i^t}M_t(p)+\varepsilon} 
\end{equation} 

Since the residual is weighted by $M_t$, unresolved but unreliable high-frequency responses contribute less to densification. The term $H_i$ is defined using the standard 3DGS screen-space gradient statistic: 
\begin{equation} 
H_i= \frac{1}{N_i}\sum_{m=1}^{N_i} \left\| \nabla_{\mathbf v_i^{xy}} \mathcal L_m \right\|_2 
\end{equation} 
where $\mathbf v_i^{xy}$ is the projected 2D position of Gaussian $G_i$ and $N_i$ is the number of valid observations accumulated for that Gaussian.

The score $S_i$ guides clone and split operations during training. Gaussians with larger $S_i$ are more likely to require additional representational support: large Gaussians with persistently high scores are preferentially split to improve local spatial resolution, while Gaussians in sparsely supported regions are preferentially cloned to increase sampling density. Because $F_i$ is modulated by $M_t$, view-inconsistent pseudo-details are less likely to trigger excessive structural growth.

In this way, densification is no longer driven only by generic image-space gradients. Instead, it is guided by a coordinated estimate of where detail is needed, where reliable spectral evidence remains unresolved, and where the current local structure is too coarse to absorb that detail.

\subsection{Overall Objective}

Combining the modules above, the full optimization objective contains a spatial-domain reconstruction term together with selective frequency-domain regularization:
\begin{equation}
L=
\lambda_{lr}L_{lr}
+\lambda_{sr}L_{sr}^{sel}
+\lambda_{amp}(t)L_{amp}
+\lambda_{ph}(t)L_{ph}
\label{loss equation}
\end{equation}

Here, $L_{lr}$ constrains the downsampled high-resolution rendering to remain consistent with the original low-resolution observations, providing stable global supervision for the scene. $L_{sr}^{sel}$ denotes the selective detail reconstruction loss weighted by the final detail-injection map $M_t$, so that super-resolved supervision is applied only where additional detail is needed and sufficiently reliable. $L_{amp}$ and $L_{ph}$ are the patch-level amplitude-spectrum and phase-spectrum losses defined above, and their corresponding weights $\lambda_{amp}(t)$ and $\lambda_{ph}(t)$ are stage-dependent. In particular, the phase term is activated later than the amplitude term, so that the model first stabilizes geometry and coarse appearance before being constrained by the more sensitive phase information.

Under this formulation, the training process is not a uniform combination of low-resolution reconstruction and super-resolved frequency matching. Instead, it follows a reliability-aware optimization strategy: geometry-guided demand estimation identifies where detail is missing, frequency-aware assessment determines whether the candidate detail is trustworthy, selective progressive regularization injects reliable spectral supervision into high-priority regions, and reliability-aware densification converts unresolved reliable detail into structural refinement of the Gaussian representation. As a result, the model is encouraged not merely to imitate super-resolved image textures, but to progressively internalize reliable details into a multi-view consistent 3D scene representation.

\section{Experiments}
\label{sec:Experiments}

\begin{figure*}[t]
\centering

\begin{minipage}[b]{0.23\textwidth}\centering {\large Mip-Splatting}\end{minipage}\hfill
\begin{minipage}[b]{0.23\textwidth}\centering {\large SplatSuRe}\end{minipage}\hfill
\begin{minipage}[b]{0.23\textwidth}\centering {\large Ours}\end{minipage}\hfill
\begin{minipage}[b]{0.23\textwidth}\centering {\large GT}\end{minipage}

\vspace{2mm}

\begin{minipage}[b]{0.23\textwidth}\centering
\includegraphics[width=\linewidth]{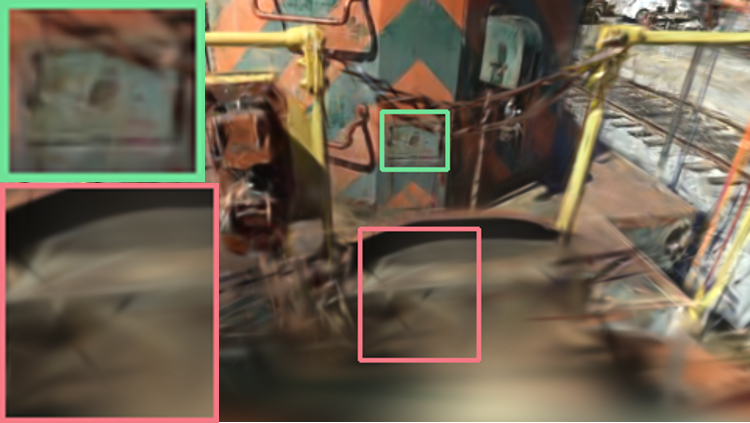}\\{\small PSNR 21.37, LPIPS 0.295}
\end{minipage}\hfill
\begin{minipage}[b]{0.23\textwidth}\centering
\includegraphics[width=\linewidth]{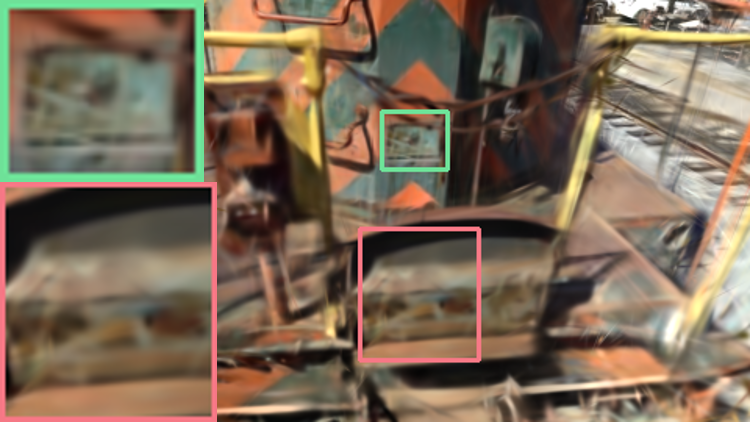}\\{\small PSNR 21.46, LPIPS 0.284}
\end{minipage}\hfill
\begin{minipage}[b]{0.23\textwidth}\centering
\includegraphics[width=\linewidth]{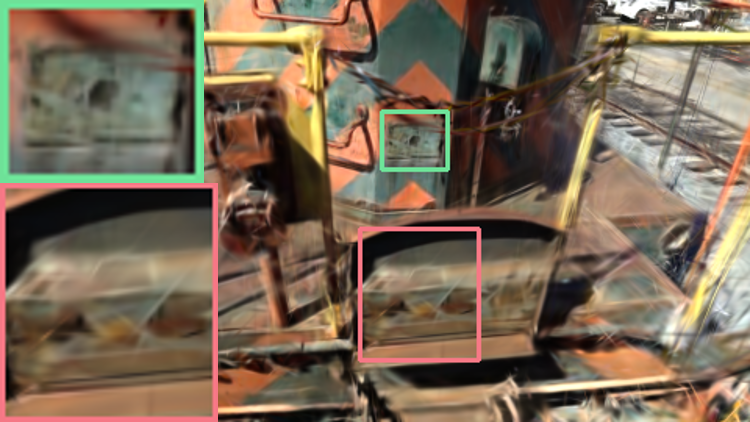}\\{\small PSNR 21.68, LPIPS 0.281}
\end{minipage}\hfill
\begin{minipage}[b]{0.23\textwidth}\centering
\includegraphics[width=\linewidth]{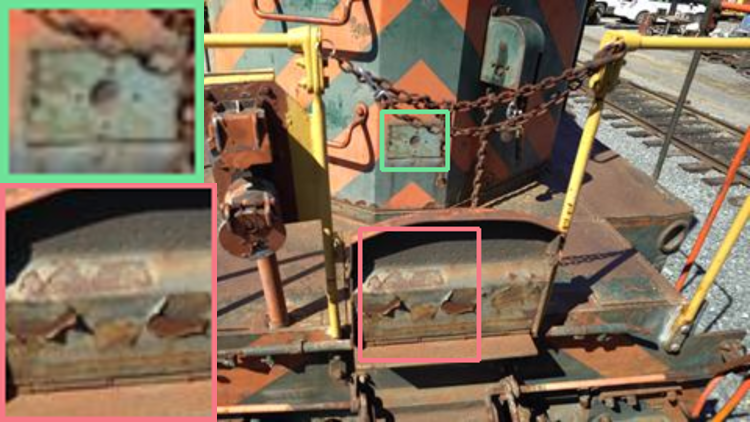}\\ \hspace{3em}
\end{minipage}

\vspace{1mm}

\begin{minipage}[b]{0.23\textwidth}\centering
\includegraphics[width=\linewidth]{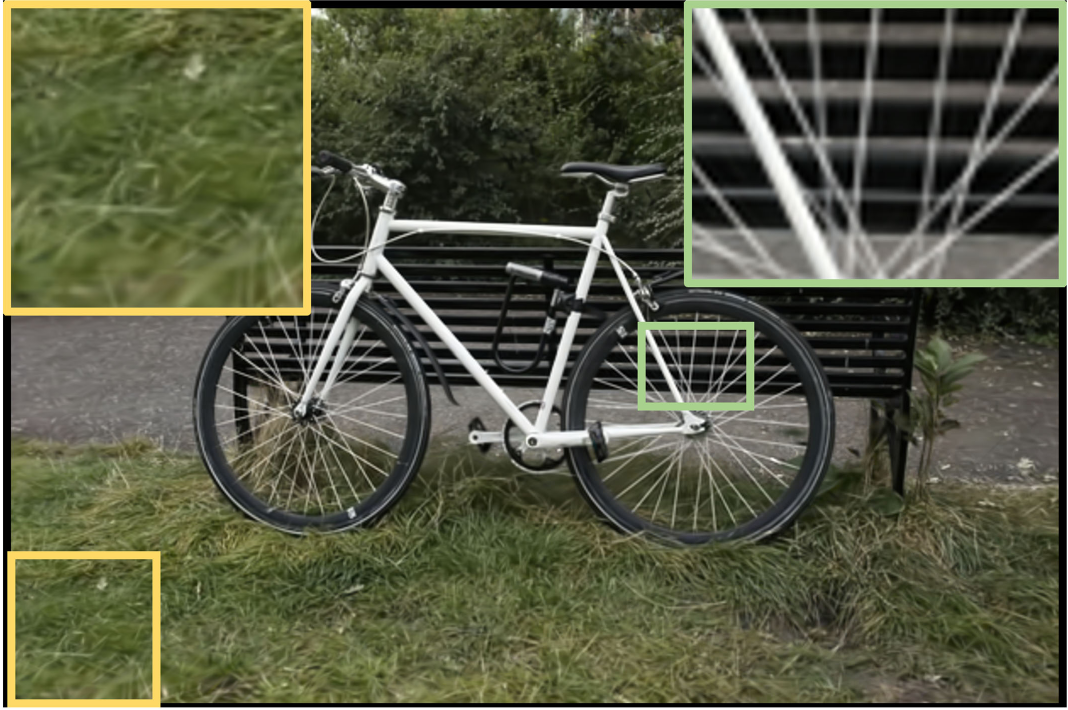}\\{\small PSNR 24.29, LPIPS 0.313}
\end{minipage}\hfill
\begin{minipage}[b]{0.23\textwidth}\centering
\includegraphics[width=\linewidth]{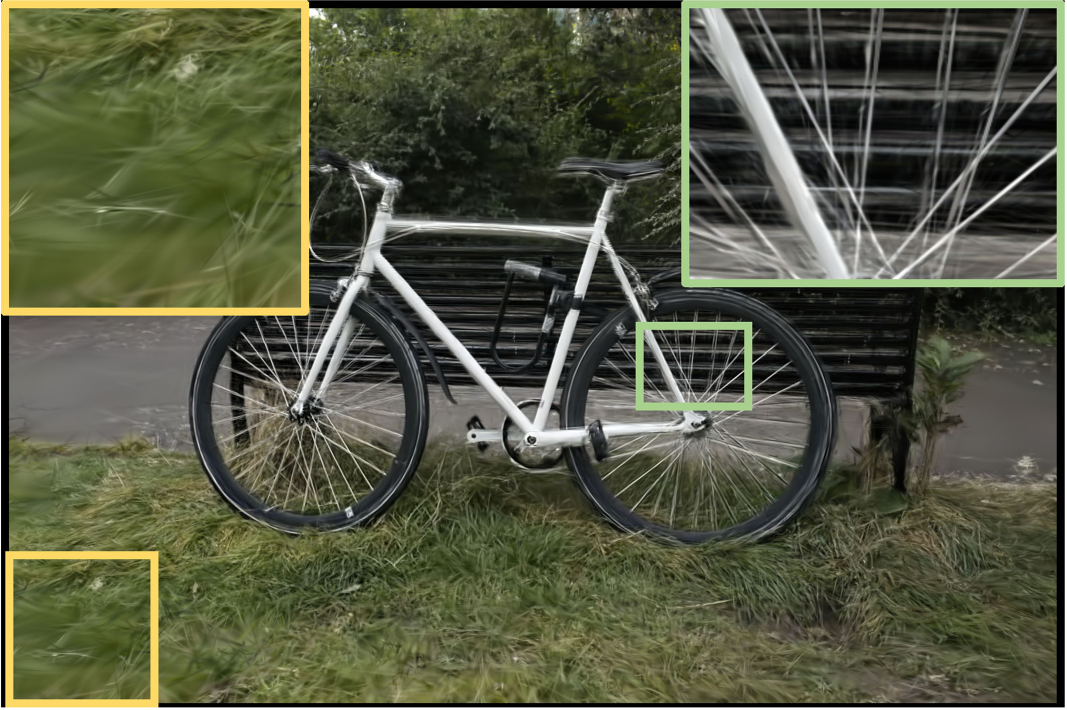}\\{\small PSNR 24.07, LPIPS 0.377}
\end{minipage}\hfill
\begin{minipage}[b]{0.23\textwidth}\centering
\includegraphics[width=\linewidth]{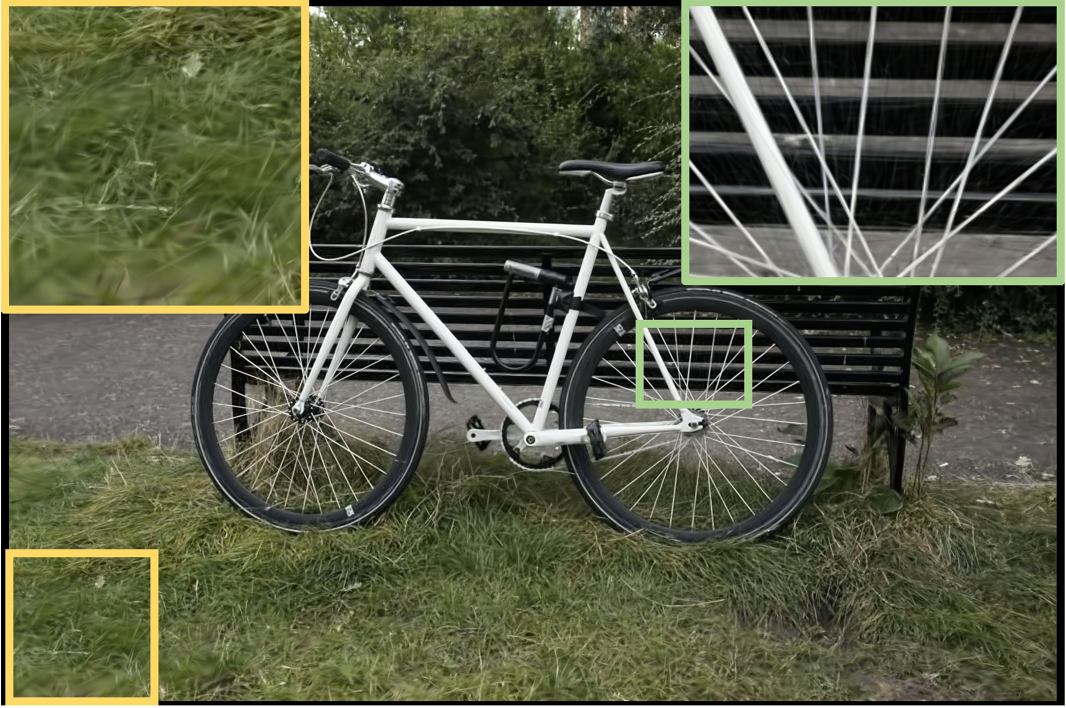}\\{\small PSNR 24.31, LPIPS 0.374}
\end{minipage}\hfill
\begin{minipage}[b]{0.23\textwidth}\centering
\includegraphics[width=\linewidth]{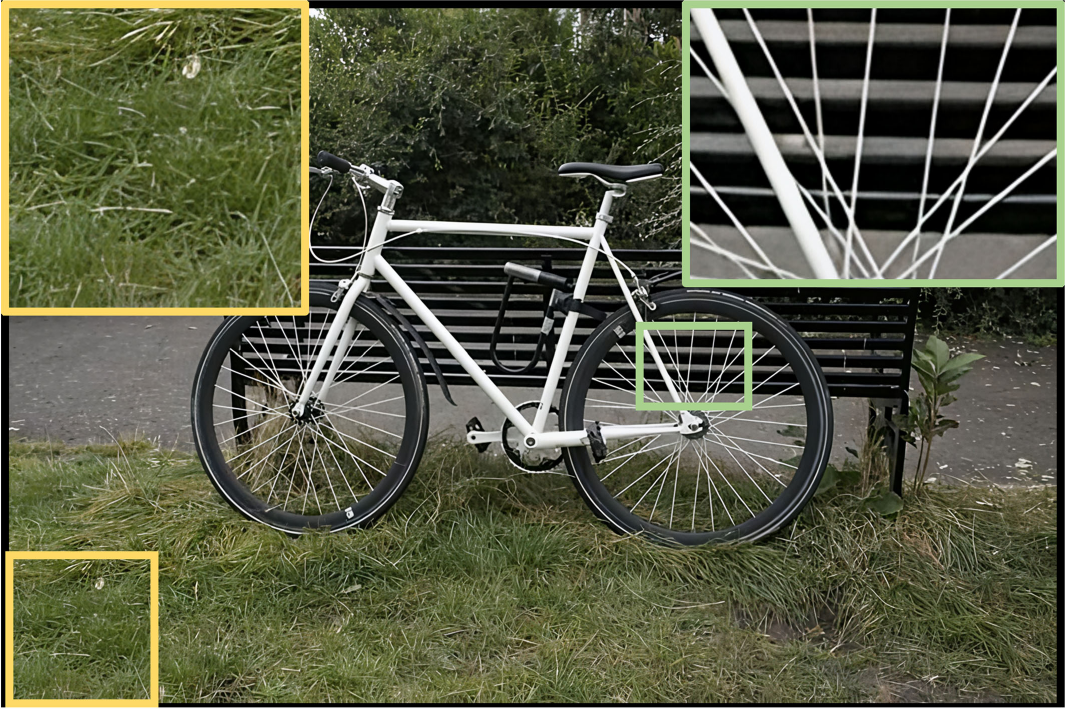}\\ \hspace{3em}
\end{minipage}

\vspace{1mm}

\begin{minipage}[b]{0.23\textwidth}\centering
\includegraphics[width=\linewidth]{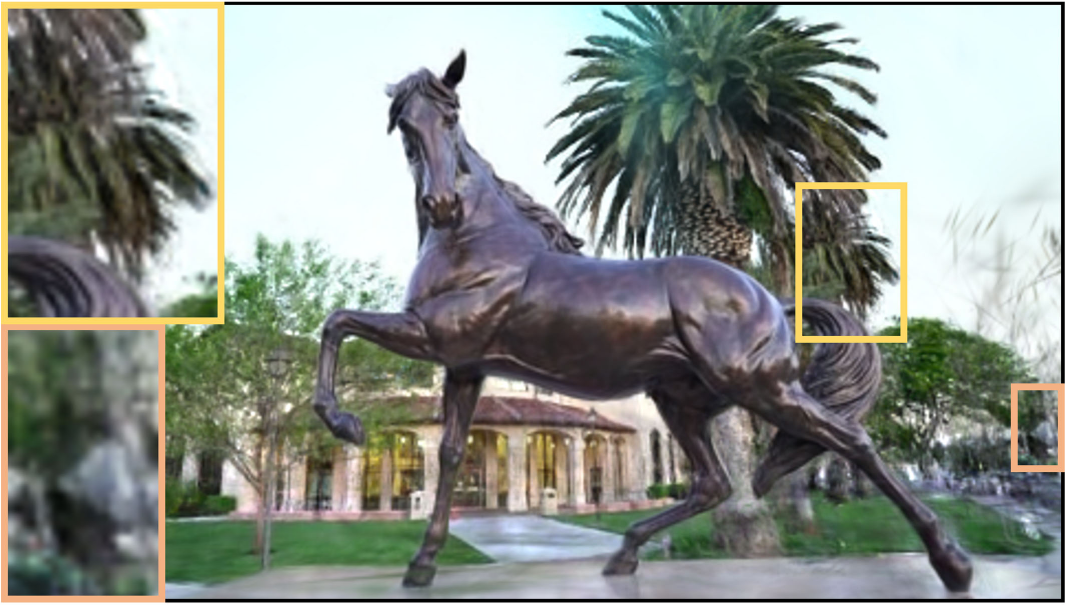}\\{\small PSNR 23.34, LPIPS 0.240}
\end{minipage}\hfill
\begin{minipage}[b]{0.23\textwidth}\centering
\includegraphics[width=\linewidth]{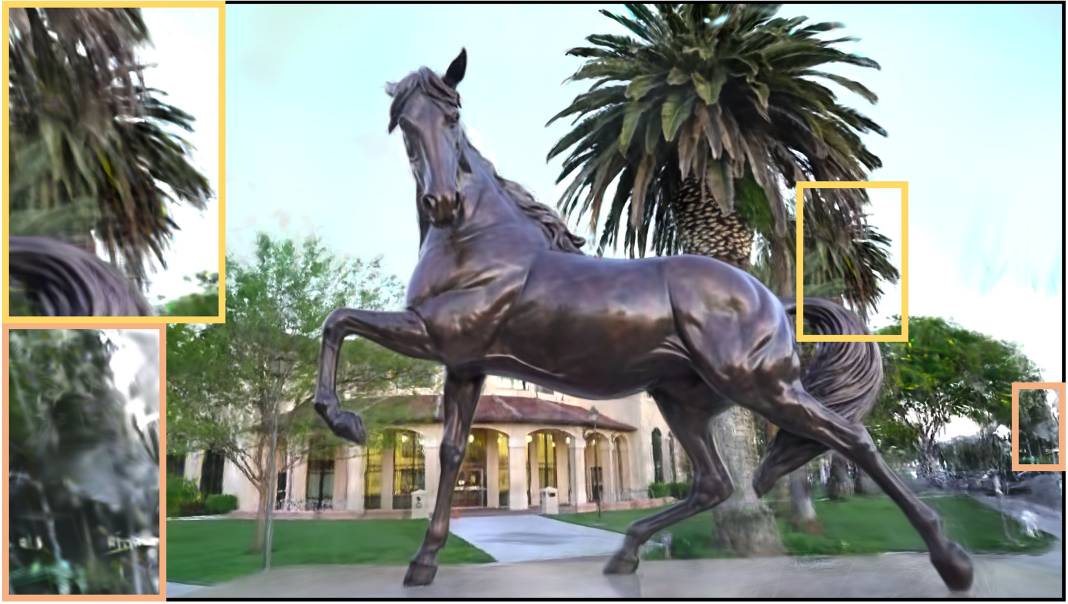}\\{\small PSNR 23.86, LPIPS 0.206}
\end{minipage}\hfill
\begin{minipage}[b]{0.23\textwidth}\centering
\includegraphics[width=\linewidth]{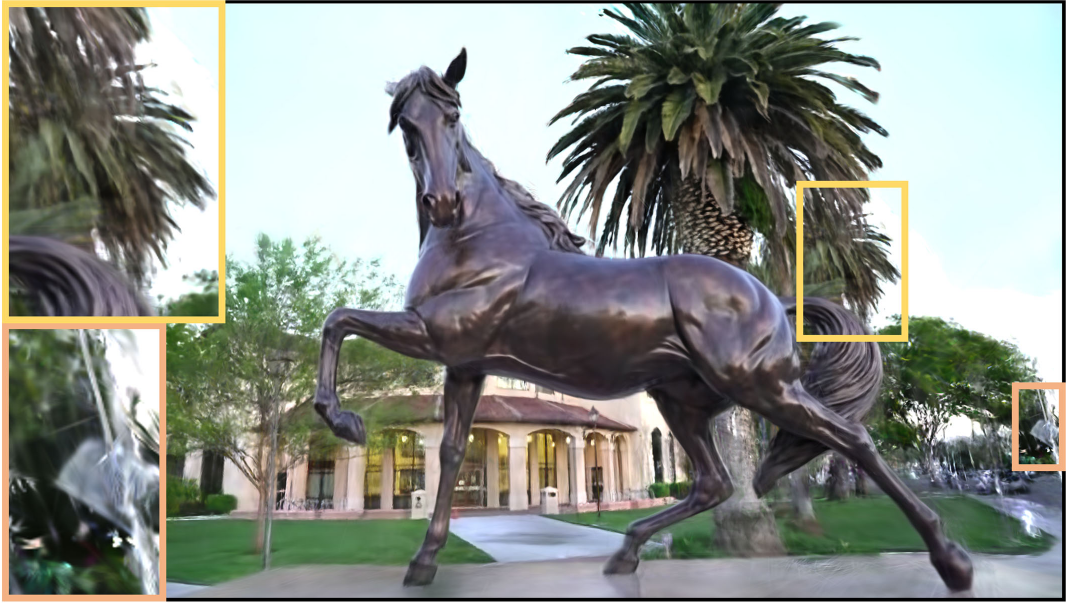}\\{\small PSNR 24.10, LPIPS 0.204}
\end{minipage}\hfill
\begin{minipage}[b]{0.23\textwidth}\centering
\includegraphics[width=\linewidth]{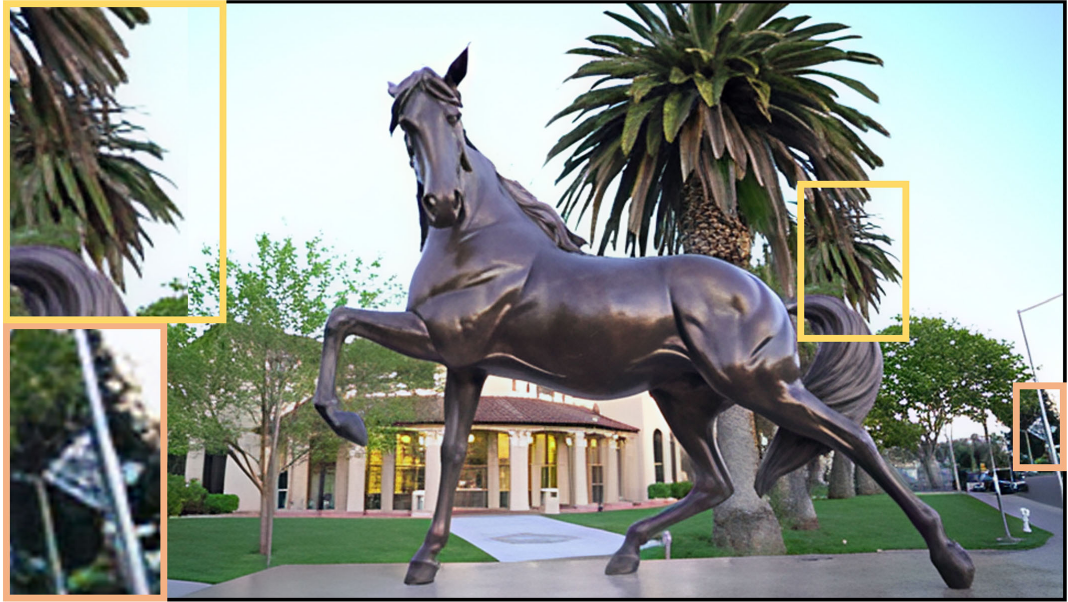}\\ \hspace{3em}
\end{minipage}

\vspace{1mm}

\begin{minipage}[b]{0.23\textwidth}\centering
\includegraphics[width=\linewidth]{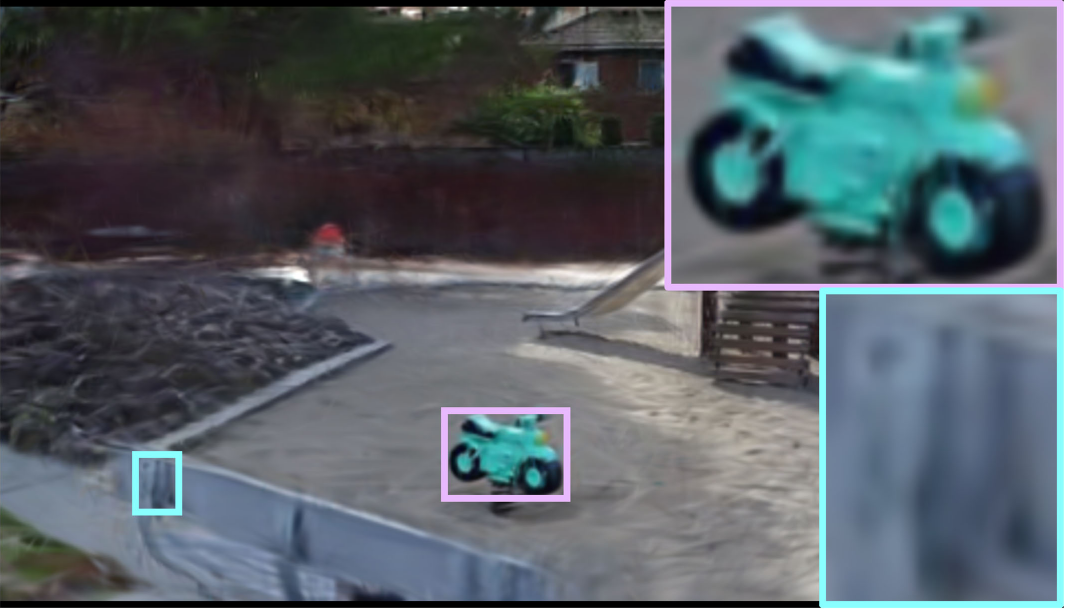}\\{\small PSNR 24.57, LPIPS 0.320}
\end{minipage}\hfill
\begin{minipage}[b]{0.23\textwidth}\centering
\includegraphics[width=\linewidth]{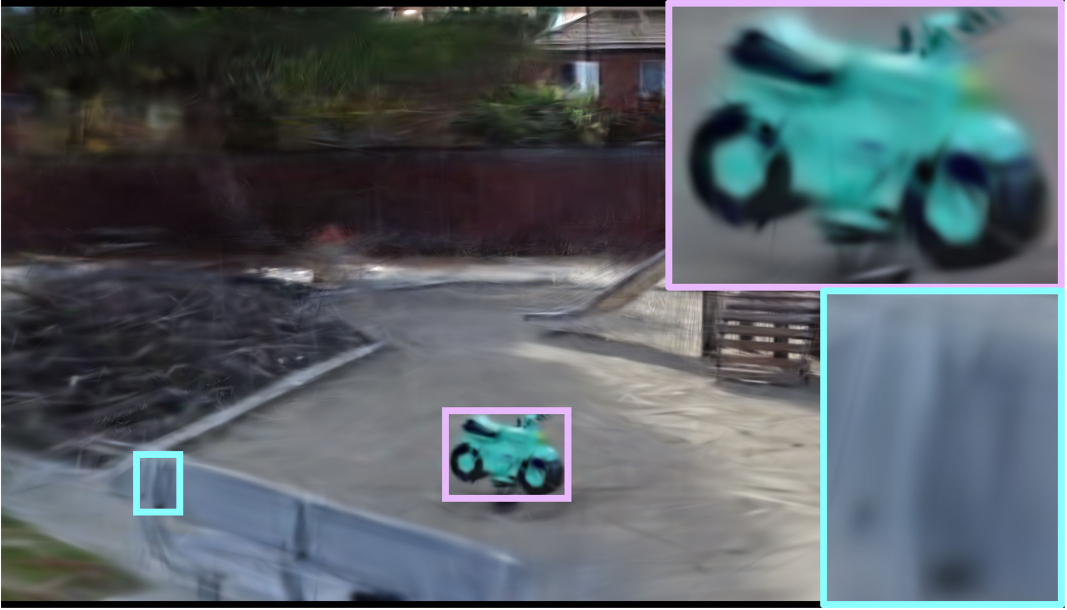}\\{\small PSNR 25.59, LPIPS 0.285}
\end{minipage}\hfill
\begin{minipage}[b]{0.23\textwidth}\centering
\includegraphics[width=\linewidth]{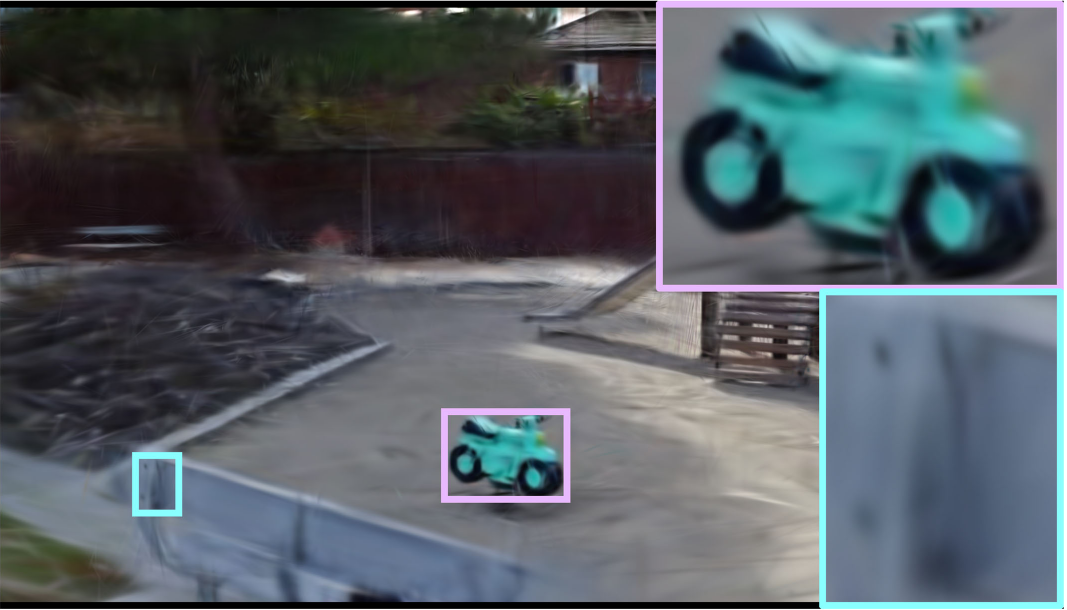}\\{\small PSNR 25.64, LPIPS 0.283}
\end{minipage}\hfill
\begin{minipage}[b]{0.23\textwidth}\centering
\includegraphics[width=\linewidth]{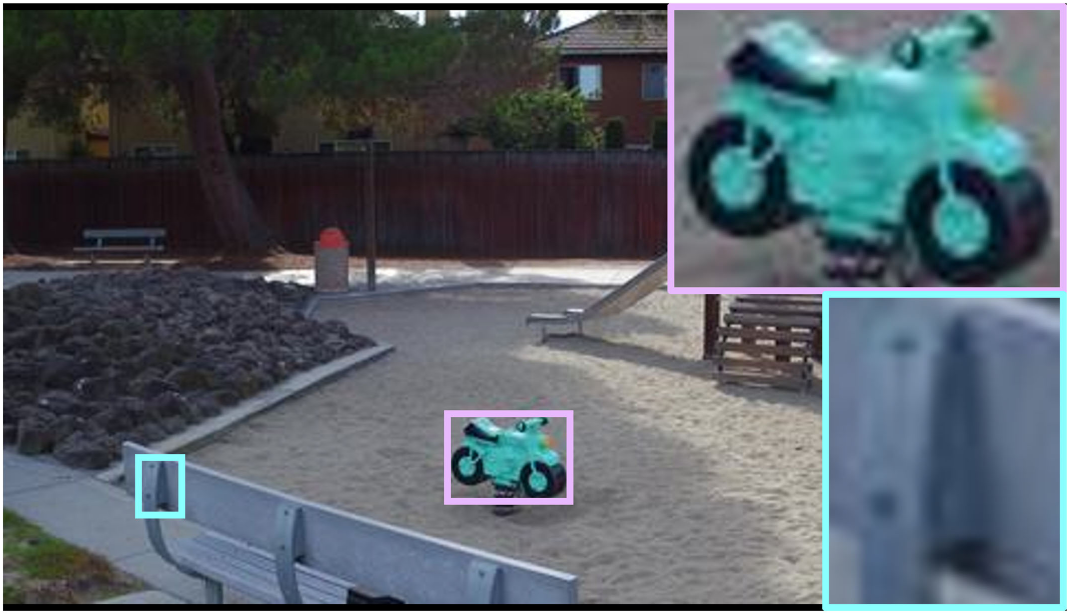}\\ \hspace{3em}
\end{minipage}

\caption{Qualitative comparisons on Tanks \& Temples and Deep Blending under 4$\times$ SR ($\tau=2$). Our method produces sharper reconstructions with fewer artifacts than Mip-Splatting and SplatSuRe.}
\label{fig:qualitative_results}
\end{figure*}

\begin{figure}[t]
\centering

\begin{minipage}[b]{0.48\linewidth}
\centering
{\small GT}\\
\vspace{1mm}
\includegraphics[width=\linewidth]{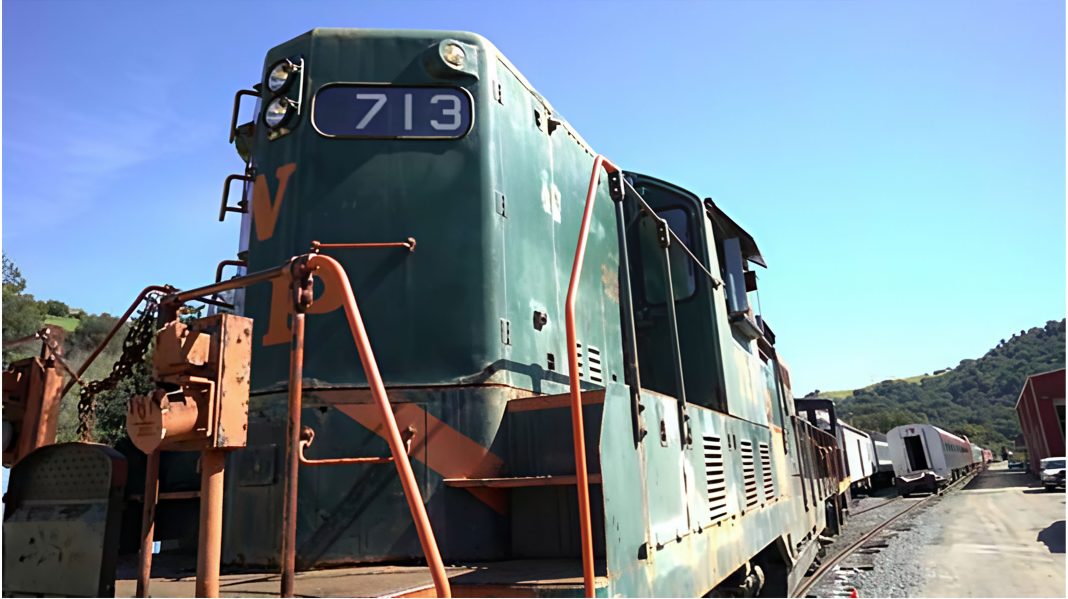}\\
\end{minipage}
\hfill
\begin{minipage}[b]{0.48\linewidth}
\centering
{\small Weight Map}\\
\vspace{1mm}
\includegraphics[width=\linewidth]{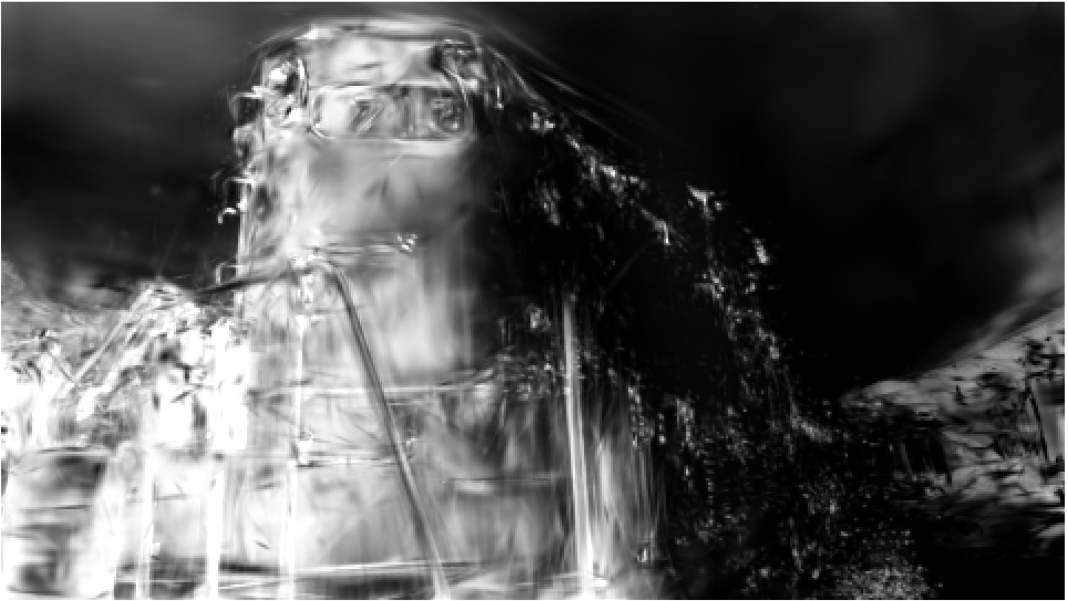}\\
\end{minipage}

\vspace{1mm}

\begin{minipage}[b]{0.48\linewidth}
\centering

\includegraphics[width=\linewidth]{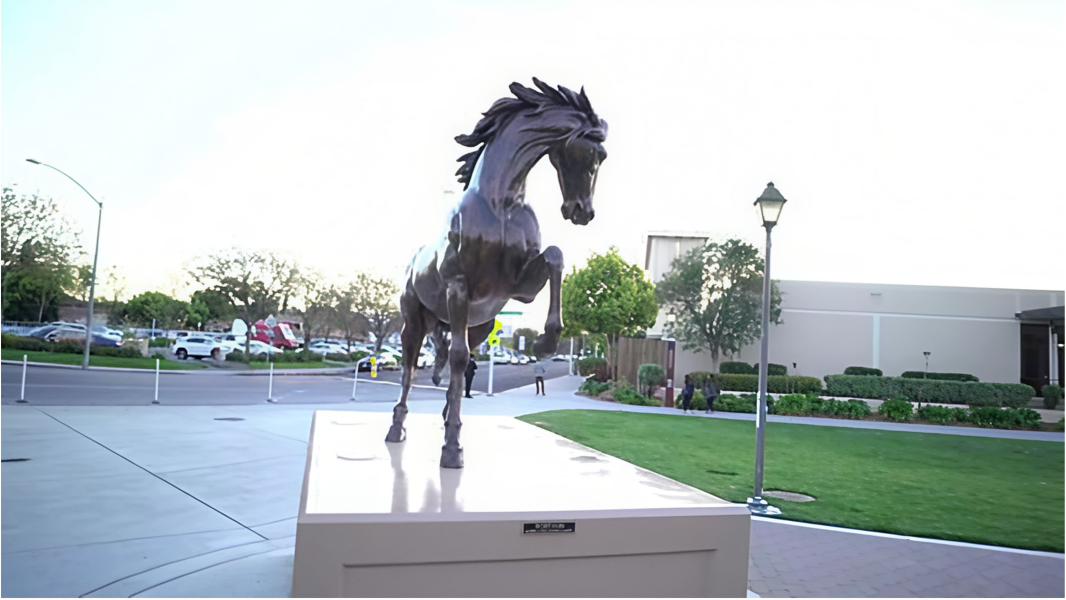}\\
\end{minipage}
\hfill
\begin{minipage}[b]{0.48\linewidth}
\centering

\includegraphics[width=\linewidth]{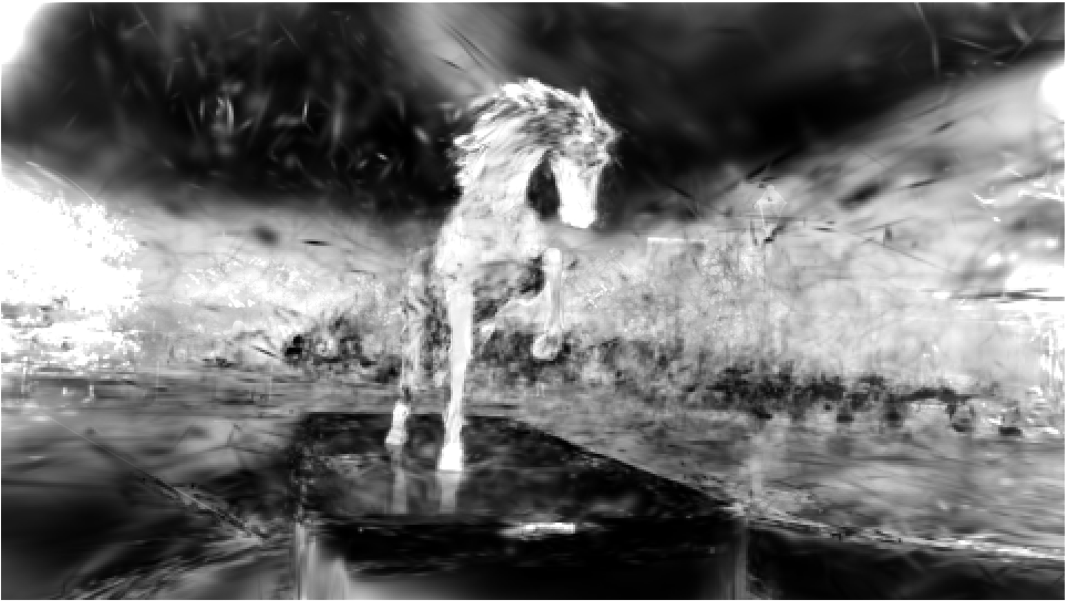}\\
\end{minipage}

\caption{Ground-truth images and detail-injection weight maps. Brighter regions indicate higher detail-injection weights.}
\label{fig:compare}
\end{figure}


Unless otherwise specified, all experiments use the same default hyperparameter setting. The values reported below are fixed across the main experiments and ablation studies rather than tuned separately for individual scenes or datasets. In the reliability assessment module, the high-pass radius used in $\mathcal{H}(\cdot)$ is set to $12.0$. For selective progressive frequency regularization, the loss weights after warm-up are $\lambda_{amp}(t)=0.05$ and $\lambda_{ph}(t)=0.001$. The amplitude branch starts from iteration $8{,}000$, the phase branch starts from iteration $12{,}000$, and both use a warm-up length of $1{,}500$ iterations. Progressive frequency annealing starts at iteration $T_0=3{,}000$ and ends at $T_d=15{,}000$, with initial frequency radius $D_0=12.0$ and maximum high-frequency weight $w_{high}^{max}=0.7$. For patch-level frequency supervision, both the patch size and FFT size are set to $64$, and at most $K=2$ patches are selected per view. The numerical stabilization constant is set to $\varepsilon=10^{-6}$. These values are used as the default configuration throughout the paper unless explicitly varied in ablation experiments.

Datasets.
We evaluate on Tanks \& Temples~\cite{knapitsch2017tanks}, Deep Blending~\cite{hedman2018deepblending}, and Mip-NeRF 360~\cite{barron2022mipnerf360}, covering indoor/outdoor scenes and varying camera trajectories. Following standard practice, every eighth image is reserved for testing. Training is performed on low-resolution inputs and evaluation at the target super-resolved resolution. We report both $4\times$ (main benchmark) and $2\times$ results. For Tanks \& Temples, we use 19 scenes (excluding two where COLMAP~\cite{schonberger2016colmap} reconstruction fails); images are downsampled to $240\times135$ and super-resolved to $960\times540$ ($4\times$) or $480\times270$ ($2\times$). For Mip-NeRF 360, nine scenes are downsampled by a factor of 4 to roughly $1000\times750$. For Deep Blending, two indoor scenes are similarly downsampled by 4 from around $1\mathrm{K}\times1\mathrm{K}$.

Baselines.
We compare against five baselines: 3DGS trained on low-resolution inputs~\cite{ref1} as the reconstruction baseline; a direct SR pipeline that enhances each training image before 3DGS optimization; Mip-Splatting~\cite{yu2024mipsplatting} for scale-aware rendering without external SR guidance; SRGS~\cite{ref3}, which applies uniform SR supervision across the full image; and SplatSuRe~\cite{asthana2025splatsure}, which selectively injects SR guidance into geometrically under-sampled regions. All methods are evaluated under the same $2\times$ and $4\times$ protocols.

Metrics.
We report reference-based metrics PSNR, SSIM, and LPIPS~\cite{zhang2018unreasonable} for pixel-level fidelity and perceptual similarity. We also report FID~\cite{heusel2017gans}, CMMD~\cite{jayasumana2024rethinkingfid}, and DreamSim~\cite{fu2023dreamsim} for perceptual distribution quality, plus no-reference metrics MUSIQ~\cite{ke2021musiq} and NIQE~\cite{mittal2013niqe} for image naturalness. This set jointly evaluates reconstruction accuracy and perceptual realism under both $2\times$ and $4\times$ settings.

\begin{table*}[!t]
\centering
\footnotesize
\renewcommand{\arraystretch}{1.12}
\setlength{\tabcolsep}{3.8pt}
\caption{Quantitative comparison on Tanks \& Temples under $4\times$ SR.}
\label{tab:main_result}
\begin{tabular*}{\textwidth}{@{\extracolsep{\fill}}p{3.2cm}cccccccc@{}}
\toprule
\multirow{2}{*}{\textbf{Method}}
& \multicolumn{8}{c}{\textbf{Tanks \& Temples~\cite{knapitsch2017tanks}}} \\
\cmidrule(lr){2-9}
& \textbf{SSIM$\uparrow$}
& \textbf{PSNR$\uparrow$}
& \textbf{LPIPS$\downarrow$}
& \textbf{FID$\downarrow$}
& \textbf{CMMD$\downarrow$}
& \textbf{DreamSim$\downarrow$}
& \textbf{MUSIQ$\uparrow$}
& \textbf{NIQE$\downarrow$} \\
\midrule
3DGS (LR) \cite{ref1}
& 0.665 & 19.41 & 0.350 & 71.88 & 1.953 & 0.0896 & \cellcolor{third}58.362 & \cellcolor{best}3.413 \\

3DGS (SR)
& 0.725 & 22.43 & 0.301 & 59.46 & 1.123 & 0.0685 & 57.080 & 4.954 \\

Mip-Splatting \cite{yu2024mipsplatting}
& 0.775 & 23.16 & 0.303 & 52.30 & 1.143 & \cellcolor{third}0.0597 & 46.602 & 5.045 \\

SRGS \cite{ref3}
& \cellcolor{third}0.776 & \cellcolor{third}23.33 & \cellcolor{third}0.286 & \cellcolor{third}49.23 & \cellcolor{third}1.041 & 0.0624 & 55.345 & 4.635 \\

SplatSuRe \cite{asthana2025splatsure}
& \cellcolor{second}0.788 & \cellcolor{second}23.84 & \cellcolor{second}0.273 & \cellcolor{second}37.80 & \cellcolor{second}1.038 & \cellcolor{second}0.0426 & \cellcolor{second}58.699 & \cellcolor{third}3.918 \\

Ours
& \cellcolor{best}0.796 & \cellcolor{best}24.08 & \cellcolor{best}0.270 & \cellcolor{best}37.42 & \cellcolor{best}1.028 & \cellcolor{best}0.0422 & \cellcolor{best}59.286 & \cellcolor{second}3.879 \\
\bottomrule
\end{tabular*}
\end{table*}

\begin{table*}[!t]
\centering
\footnotesize
\setlength{\tabcolsep}{3.8pt}
\renewcommand{\arraystretch}{1.15}

\caption{Quantitative comparison on Deep Blending~\cite{hedman2018deepblending} and Mip-NeRF 360~\cite{barron2022mipnerf360} under $4\times$ SR ($\tau=2$).}
\label{tab:dataset_compare}

\begin{tabular*}{\textwidth}{@{\extracolsep{\fill}}p{3.2cm}cccccccccc@{}}
\toprule
\multirow{2}{*}{\textbf{Method}}
& \multicolumn{5}{c}{\textbf{Deep Blending~\cite{hedman2018deepblending}}}
& \multicolumn{5}{c}{\textbf{Mip-NeRF 360~\cite{barron2022mipnerf360}}} \\
\cmidrule(lr){2-6} \cmidrule(lr){7-11}
& \textbf{SSIM$\uparrow$}
& \textbf{PSNR$\uparrow$}
& \textbf{LPIPS$\downarrow$}
& \textbf{CMMD$\downarrow$}
& \textbf{DreamSim$\downarrow$}
& \textbf{SSIM$\uparrow$}
& \textbf{PSNR$\uparrow$}
& \textbf{LPIPS$\downarrow$}
& \textbf{CMMD$\downarrow$}
& \textbf{DreamSim$\downarrow$} \\
\midrule
3DGS (LR) \cite{ref1}
& 0.844 & 26.14 & 0.340 & 0.861 & 0.0576
& 0.639 & 20.65 & 0.388 & 0.648 & 0.0520 \\

3DGS (SR)
& 0.855 & 27.10 & 0.323 & 0.748 & 0.0503
& 0.701 & 24.32 & 0.355 & 0.719 & 0.0366 \\

Mip-Splatting \cite{yu2024mipsplatting}
& \cellcolor{third}0.861 & \cellcolor{third}28.54 & 0.327 & 0.690 & \cellcolor{third}0.0405
& \cellcolor{best}0.763 & \cellcolor{second}26.61 & \cellcolor{best}0.291 & \cellcolor{best}0.183 & \cellcolor{best}0.0124 \\

SRGS \cite{ref3}
& 0.856 & 27.85 & \cellcolor{third}0.314 & \cellcolor{third}0.637 & 0.0408
& 0.735 & 25.92 & \cellcolor{second}0.317 & \cellcolor{third}0.341 & 0.0194 \\

SplatSuRe \cite{asthana2025splatsure}
& \cellcolor{second}0.871 & \cellcolor{second}29.33 & \cellcolor{second}0.310 & \cellcolor{second}0.494 & \cellcolor{second}0.0330
& \cellcolor{third}0.744 & \cellcolor{third}26.48 & 0.325 & 0.342 & \cellcolor{third}0.0179 \\

Ours
& \cellcolor{best}0.879 & \cellcolor{best}29.62 & \cellcolor{best}0.307 & \cellcolor{best}0.489 & \cellcolor{best}0.0329
& \cellcolor{second}0.752 & \cellcolor{best}26.74 & \cellcolor{third}0.321 & \cellcolor{second}0.339 & \cellcolor{second}0.0178 \\
\bottomrule
\end{tabular*}
\end{table*}
\begin{table*}[!t]
\centering
\footnotesize
\renewcommand{\arraystretch}{1.12}
\setlength{\tabcolsep}{3.8pt}
\caption{Quantitative comparison on Tanks \& Temples~\cite{knapitsch2017tanks} under $2\times$ SR ($\tau=2$).}
\label{tab:main_result2}

\begin{tabular*}{\textwidth}{@{\extracolsep{\fill}}p{3.6cm}cccccccc@{}}
\toprule
\multirow{2}{*}{\textbf{Method}}
& \multicolumn{8}{c}{\textbf{Tanks \& Temples~\cite{knapitsch2017tanks}}} \\
\cmidrule(lr){2-9}
& \textbf{SSIM$\uparrow$}
& \textbf{PSNR$\uparrow$}
& \textbf{LPIPS$\downarrow$}
& \textbf{FID$\downarrow$}
& \textbf{CMMD$\downarrow$}
& \textbf{DreamSim$\downarrow$}
& \textbf{MUSIQ$\uparrow$}
& \textbf{NIQE$\downarrow$} \\
\midrule
3DGS (LR) \cite{ref1}
& 0.665 & 19.41 & 0.350 & 71.88 & 1.953 & 0.0896 & \cellcolor{third}58.362 & \cellcolor{best}3.413 \\

3DGS (SR)
& 0.709 & 20.92 & 0.325 & 68.12 & 1.544 & 0.0812 & 53.639 & 5.449 \\

Mip-Splatting \cite{yu2024mipsplatting}
& 0.776 & 23.16 & 0.303 & 52.30 & 1.143 & \cellcolor{third}0.0597 & 46.602 & 5.045 \\

SRGS \cite{ref3}
& \cellcolor{third}0.721 & \cellcolor{third}21.37 & \cellcolor{third}0.318 & \cellcolor{third}61.48 & \cellcolor{third}1.497 & 0.0784 & 50.678 & 5.099 \\

SplatSuRe \cite{asthana2025splatsure}
& \cellcolor{second}0.775 & \cellcolor{second}23.23 & \cellcolor{second}0.300 & \cellcolor{second}48.94 & \cellcolor{second}1.122 & \cellcolor{second}0.0541 & \cellcolor{second}58.953 & \cellcolor{third}3.379 \\

Ours
& \cellcolor{best}0.782 & \cellcolor{best}23.44 & \cellcolor{best}0.298 & \cellcolor{best}47.20 & \cellcolor{best}1.081 & \cellcolor{best}0.0517 & \cellcolor{best}59.537 & \cellcolor{second}3.271 \\
\bottomrule
\end{tabular*}
\end{table*}
\begin{table*}[!t]
\centering
\footnotesize
\setlength{\tabcolsep}{3.8pt}
\renewcommand{\arraystretch}{1.15}

\caption{Quantitative comparison on Deep Blending~\cite{hedman2018deepblending} and Mip-NeRF 360~\cite{barron2022mipnerf360} under $2\times$ SR ($\tau=2$).}
\label{tab:dataset_compare2}

\begin{tabular*}{\textwidth}{@{\extracolsep{\fill}}p{3.2cm}cccccccccc@{}}
\toprule
\multirow{2}{*}{\textbf{Method}}
& \multicolumn{5}{c}{\textbf{Deep Blending~\cite{hedman2018deepblending}}}
& \multicolumn{5}{c}{\textbf{Mip-NeRF 360~\cite{barron2022mipnerf360}}} \\
\cmidrule(lr){2-6} \cmidrule(lr){7-11}
& \textbf{SSIM$\uparrow$}
& \textbf{PSNR$\uparrow$}
& \textbf{LPIPS$\downarrow$}
& \textbf{CMMD$\downarrow$}
& \textbf{DreamSim$\downarrow$}
& \textbf{SSIM$\uparrow$}
& \textbf{PSNR$\uparrow$}
& \textbf{LPIPS$\downarrow$}
& \textbf{CMMD$\downarrow$}
& \textbf{DreamSim$\downarrow$} \\
\midrule
3DGS (LR) \cite{ref1}
& 0.844 & 26.14 & 0.340 & 0.861 & 0.0576
& 0.639 & 20.65 & 0.388 & 0.648 & 0.0520 \\

3DGS (SR)
& 0.850 & 26.62 & 0.332 & 0.805 & 0.0540
& 0.670 & 22.49 & 0.371 & 0.788 & 0.0456 \\

Mip-Splatting \cite{yu2024mipsplatting}
& \cellcolor{third}0.861 & \cellcolor{second}28.54 & \cellcolor{third}0.327 & \cellcolor{third}0.690 & \cellcolor{third}0.0405
& \cellcolor{best}0.763 & \cellcolor{best}26.61 & \cellcolor{best}0.291 & \cellcolor{best}0.183 & \cellcolor{best}0.0124 \\

SRGS \cite{ref3}
& 0.850 & 27.00 & \cellcolor{third}0.327 & 0.749 & 0.0492
& 0.687 & 23.27 & \cellcolor{second}0.352 & 0.505 & 0.0357 \\

SplatSuRe \cite{asthana2025splatsure}
& \cellcolor{second}0.862 & \cellcolor{third}28.45 & \cellcolor{second}0.320 & \cellcolor{second}0.684 & \cellcolor{second}0.0396
& \cellcolor{third}0.691 & \cellcolor{third}23.57 & 0.356 & \cellcolor{third}0.495 & \cellcolor{third}0.0350 \\

Ours
& \cellcolor{best}0.863 & \cellcolor{best}28.59 & \cellcolor{best}0.319 & \cellcolor{best}0.679 & \cellcolor{best}0.0391
& \cellcolor{second}0.695 & \cellcolor{second}23.70 & \cellcolor{third}0.354 & \cellcolor{second}0.493 & \cellcolor{second}0.0349 \\
\bottomrule
\end{tabular*}
\end{table*}

Training time.
We also report the average post-map optimization time per scene under both 2$\times$ and 4$\times$ settings. All experiments are conducted on a single NVIDIA RTX TITAN GPU and trained for 20,000 iterations. The reported time is measured from the training stage after the initial detail-injection weight maps are obtained, excluding COLMAP preprocessing, SR reference generation, and initial weight-map construction. As shown in Table~\ref{tab:training_time}, the 4$\times$ setting consistently requires more optimization time than the 2$\times$ setting. This is expected because the higher target resolution increases the cost of high-resolution rendering, reliability-aware frequency supervision, and patch-level Fourier operations. Nevertheless, the additional cost remains moderate under the default configuration, showing that the proposed reliability-aware detail injection can be applied with practical training overhead.

\begin{table}[t]
\centering
\footnotesize
\caption{Average post-map optimization time per scene.}
\label{tab:training_time}
\setlength{\tabcolsep}{6pt}
\renewcommand{\arraystretch}{1.12}
\begin{tabular*}{\columnwidth}{@{\hspace{0.8em}\extracolsep{\fill}}lccc@{\hspace{0.8em}}}
\hline
Dataset & 2$\times$ Time & 4$\times$ Time & Increase \\
\hline
Tanks \& Temples & 39 min & 46 min & 17.9\% \\
Mip-NeRF 360 & 76 min & 93 min & 22.4\% \\
Deep Blending & 63 min & 81 min & 28.6\% \\
\hline
\end{tabular*}
\end{table}

Qualitative results.
Figure~\ref{fig:qualitative_results} compares our method with Mip-Splatting and SplatSuRe on Tanks \& Temples and Deep Blending under the $4\times$ setting. Our method produces sharper reconstructions around thin boundaries, small distant objects, and textured areas, while avoiding excessive high-frequency artifacts. Figure~\ref{fig:compare} visualizes the corresponding detail-injection weight maps: brighter regions indicate higher weights concentrated around structurally complex areas, showing that the proposed strategy adaptively assigns stronger supervision to reliable under-detailed regions rather than uniformly enhancing the whole image.

Quantitative results.
Tables~\ref{tab:main_result} and~\ref{tab:dataset_compare} report the $4\times$ results, while Tables~\ref{tab:main_result2} and~\ref{tab:dataset_compare2} report the $2\times$ results. Under $4\times$, our method achieves the best overall performance on Tanks \& Temples (best on all metrics except NIQE) and Deep Blending (best SSIM, PSNR, LPIPS, CMMD, and DreamSim). On Mip-NeRF 360, our method achieves the best PSNR and remains competitive on SSIM, CMMD, and DreamSim. The $2\times$ results show similar trends, confirming that the proposed reliability-aware detail injection generalizes across different upsampling scales.

\section{Ablation Study}
\label{sec:Ablation Study}

We evaluate the main design choices through ablation experiments, examining the contribution of each component, the effect of geometry-guided demand estimation, selective progressive frequency regularization, reliability assessment design, densification design, and the number of selected patches. All ablations follow the same training schedule and evaluation protocol as the main experiments.

\begin{table}[t]
\centering
\footnotesize
\caption{Progressive ablation on the bicycle scene (Mip-NeRF 360).}
\label{tab:ablation_module}
\setlength{\tabcolsep}{6pt}
\renewcommand{\arraystretch}{1.12}
\begin{tabular*}{\columnwidth}{@{\hspace{0.8em}\extracolsep{\fill}}lccc@{\hspace{0.8em}}}
\hline
Variant & PSNR$\uparrow$ & LPIPS$\downarrow$ & SSIM$\uparrow$ \\
\hline
w/o Reliability assessment & 21.75 & 0.273 & 0.763 \\
w/o Frequency regularization & 21.87 & 0.285 & 0.771 \\
w/o Gaussian densification  & 20.17 & 0.291 & 0.681 \\
Full & \textbf{22.56} & \textbf{0.251} & \textbf{0.797} \\
\hline
\end{tabular*}
\end{table}

\subsection{Effect of Progressive Reliability-Aware Component Integration}

Table~\ref{tab:ablation_module} reports component-wise ablations on the bicycle scene from Mip-NeRF 360. The full model achieves the best performance (PSNR 22.56, LPIPS 0.251, SSIM 0.797). Removing reliability assessment degrades results (PSNR 21.75, LPIPS 0.273, SSIM 0.763), showing its importance for filtering unreliable high-frequency details. Removing progressive frequency regularization also hurts performance (PSNR 21.87, LPIPS 0.285, SSIM 0.771), indicating that high-frequency supervision should be introduced gradually. The largest drop occurs without reliability-aware densification (PSNR 20.17, LPIPS 0.291, SSIM 0.681), confirming that loss-based supervision alone is insufficient without adequate Gaussian capacity. These components play complementary roles.

The complementarity among these components can be understood through their distinct roles in the optimization pipeline. Reliability assessment acts as a gatekeeper that prevents unreliable high-frequency content from entering the 3D representation, thereby preserving multi-view consistency. Progressive frequency regularization provides a curriculum-like training strategy that gradually increases the complexity of frequency constraints, allowing the Gaussian parameters to adapt structurally before finer details are enforced. Reliability-aware densification addresses the representation capacity bottleneck by strategically increasing Gaussian primitives in regions where the loss signal indicates both demand and reliability. When any of these mechanisms is missing, the system compensates imperfectly: without reliability filtering, the optimizer absorbs view-inconsistent details; without progressive scheduling, high-frequency constraints destabilize early optimization; and without targeted densification, existing Gaussians over-stretch to cover under-resolved regions, producing blurry or artifact-prone reconstructions.

\subsection{Effect of Geometry-Guided Detail-Demand Estimation}

\begin{figure}[!t]
\centering
\includegraphics[width=3in]{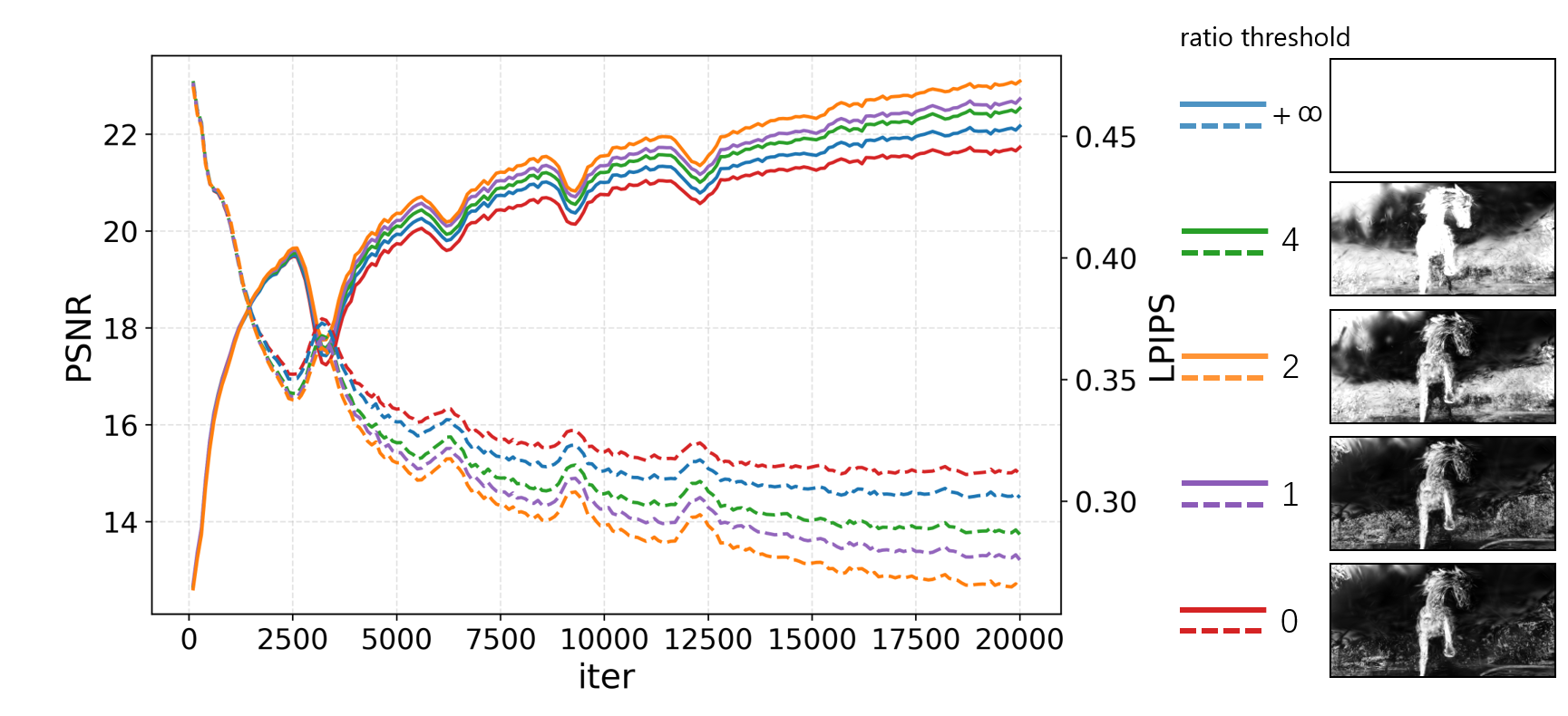}
\caption{Effect of threshold $\tau$ on PSNR and LPIPS (horse scene, Tanks \& Temples).}
\label{fig_ablation1}
\end{figure}

We evaluate the threshold parameter $\tau$ on the horse scene from Tanks \& Temples. As shown in Figure~\ref{fig_ablation1}, $\tau=2$ achieves the best balance: smaller values broaden detail injection to regions that do not benefit, while larger values overly restrict supervision in genuinely under-detailed areas. A moderate selectivity provides the most suitable spatial range for reliable detail injection.

This threshold-driven behavior reflects the underlying cross-view sampling geometry. When $\tau$ is small, the detail-injection map becomes overly conservative, limiting supervision to only the most severely under-sampled regions and leaving many areas that could benefit from enhancement unsupervised. Conversely, when $\tau$ is large, the map expands to include regions that are already reasonably well-sampled in at least some views, causing redundant SR supervision that may introduce conflicting details across different camera positions. The optimal $\tau=2$ strikes a balance by targeting regions where the aggregate sampling evidence genuinely indicates insufficient detail, without overextending into areas where existing observations already provide adequate structural information. This confirms that geometry-guided demand estimation is not merely a binary mask but a calibrated spatial prior that narrows the scope of detail injection to the most promising candidate regions.

\subsection{Effect of Selective Progressive Frequency Regularization}

\begin{figure}[!t]
\centering
\includegraphics[width=3in]{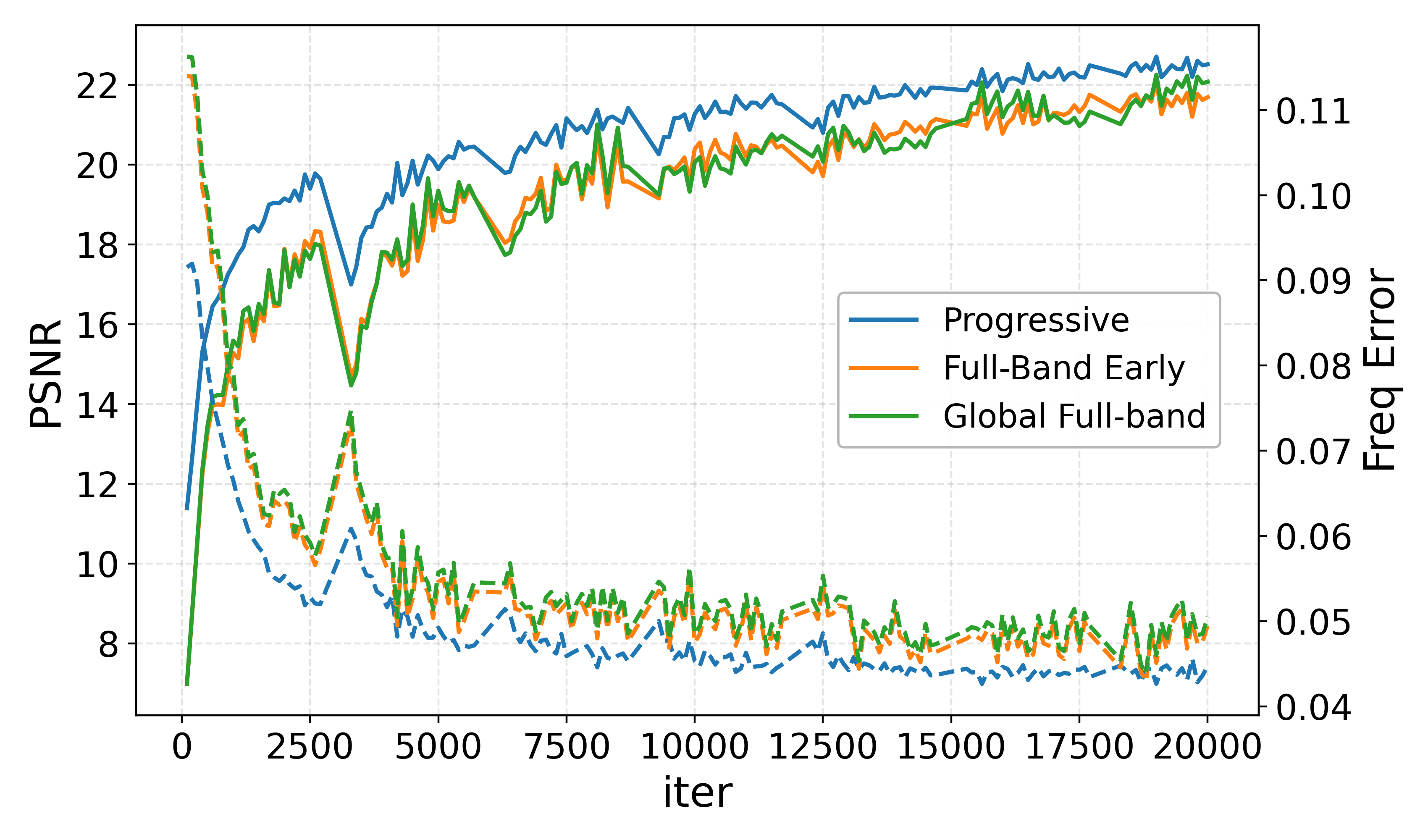}
\caption{Comparison of frequency regularization variants (horse scene, Tanks \& Temples). Our selective progressive design achieves more stable optimization and better image quality.}
\label{fig_ablation2}
\end{figure}

We compare the full model with two variants on the horse scene from Tanks \& Temples: one without the progressive schedule (full-band constraints from the start) and one without spatial selectivity (global frequency supervision). Figure~\ref{fig_ablation2} shows that both variants underperform the full model. Early full-band constraints destabilize optimization, while global supervision introduces unreliable details. This confirms that the detail-injection map should control where frequency supervision is applied and the progressive schedule should control when finer constraints are introduced.

The interaction between spatial selectivity and temporal progression is particularly important for low-resolution 3DGS. Spatial selectivity ensures that frequency supervision is applied only where the detail-injection map indicates both demand and reliability, preventing the optimizer from wasting capacity on regions where high-frequency content is either unnecessary or untrustworthy. Temporal progression, meanwhile, prevents premature enforcement of fine-grained spectral constraints before the Gaussian primitives have established reasonable geometric and photometric foundations. By coordinating these two controls, the full model avoids the instability that often plagues early-phase frequency regularization while still achieving detailed reconstruction in later iterations. This suggests that effective frequency-domain supervision in 3DGS requires not only appropriate loss formulations but also careful orchestration of where and when those losses are activated.

\subsection{Ablation on Frequency-Aware Reliability Assessment}

We analyze the three cues in the reliability map on the train scene from Tanks \& Temples: $E_t$ (structural support), $G_t$ (unresolved high-frequency evidence), and $X_t$ (cross-view inconsistency penalty).

\begin{figure*}[t]
\centering

\begin{minipage}[b]{0.33\textwidth}\centering {w/o $E_t$}\end{minipage}\hfill
\begin{minipage}[b]{0.33\textwidth}\centering {FULL}\end{minipage}\hfill
\begin{minipage}[b]{0.33\textwidth}\centering {GT}\end{minipage}

\vspace{1mm}

\begin{minipage}[b]{0.33\textwidth}\centering
\includegraphics[width=\linewidth]{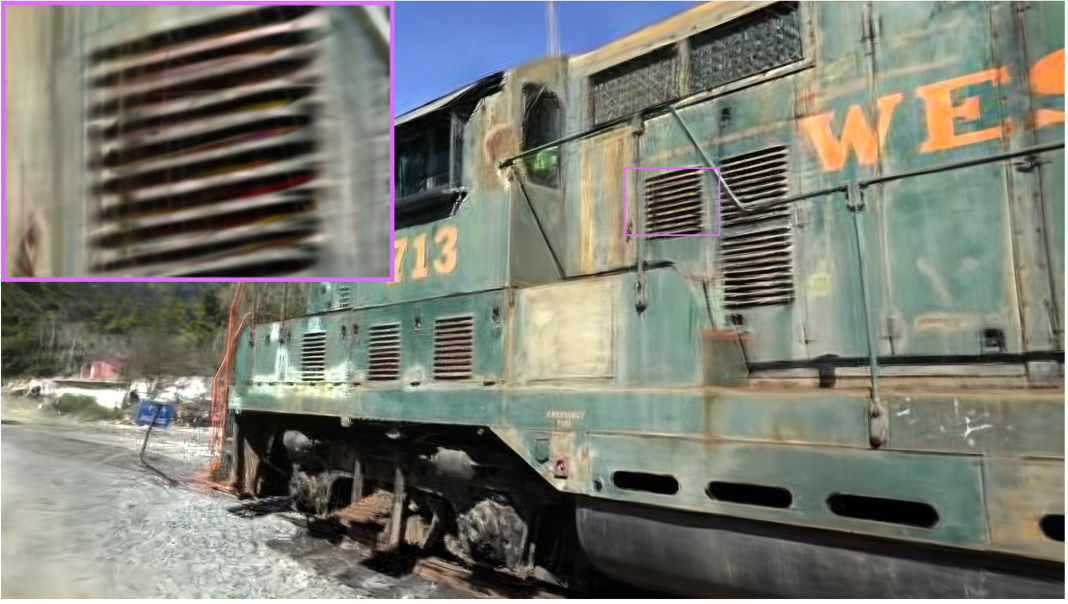}
\end{minipage}\hfill
\begin{minipage}[b]{0.33\textwidth}\centering
\includegraphics[width=\linewidth]{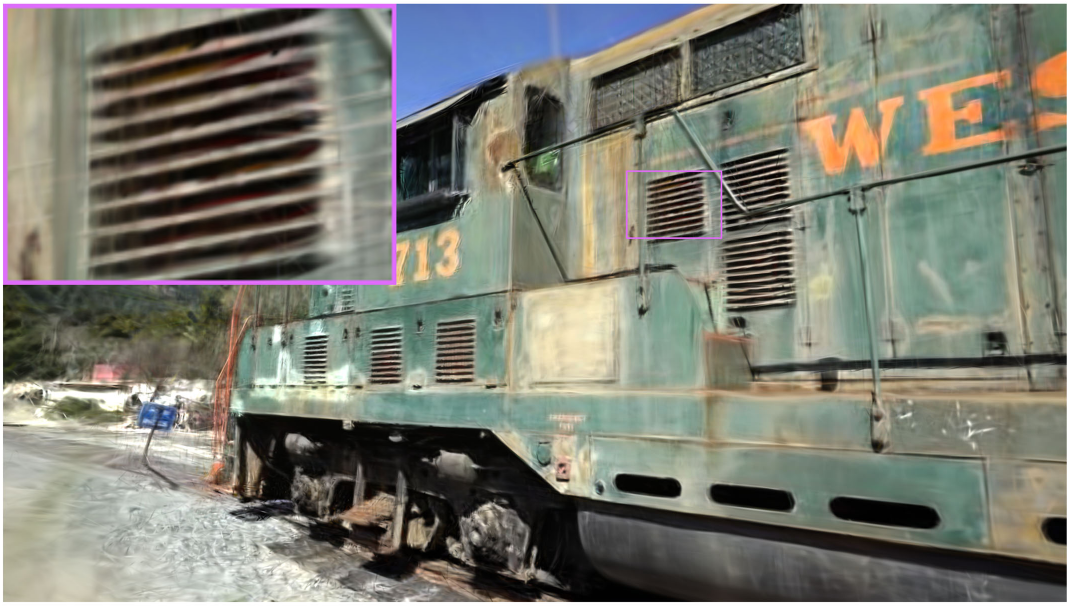}
\end{minipage}\hfill
\begin{minipage}[b]{0.33\textwidth}\centering
\includegraphics[width=\linewidth]{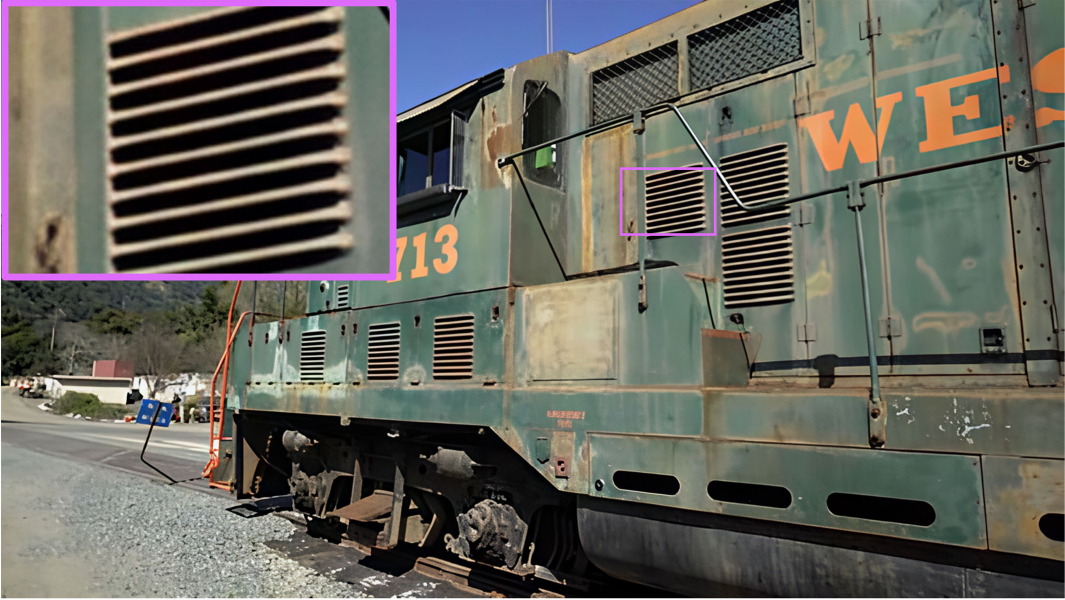}
\end{minipage}

\vspace{1mm}

\begin{minipage}[b]{0.33\textwidth}\centering {w/o $G_t$}\end{minipage}\hfill
\begin{minipage}[b]{0.33\textwidth}\centering {FULL}\end{minipage}\hfill
\begin{minipage}[b]{0.33\textwidth}\centering {GT}\end{minipage}

\vspace{1mm}

\begin{minipage}[b]{0.33\textwidth}\centering
\includegraphics[width=\linewidth]{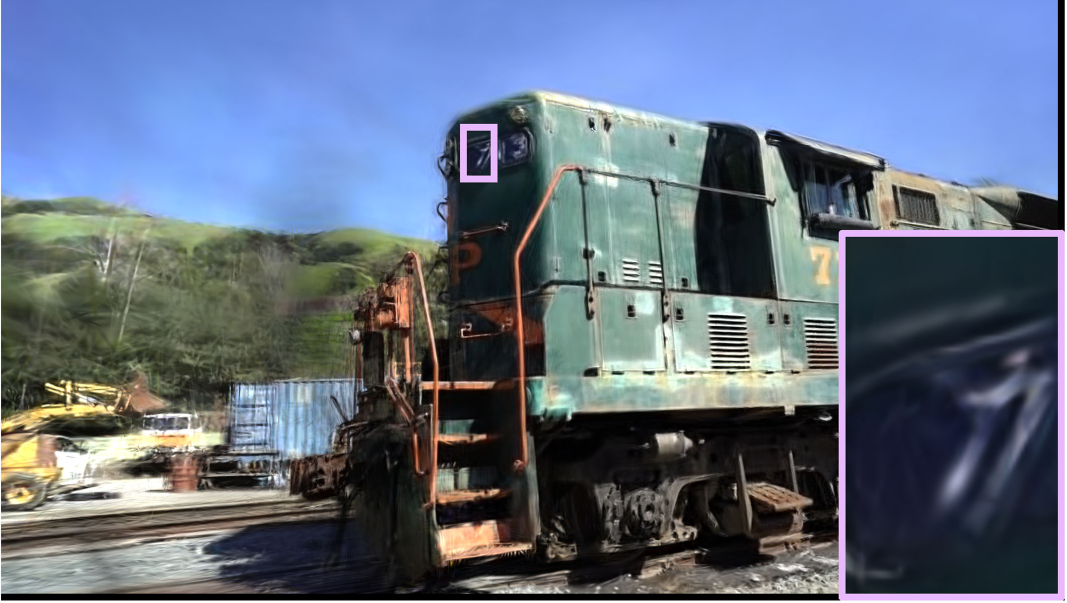}
\end{minipage}\hfill
\begin{minipage}[b]{0.33\textwidth}\centering
\includegraphics[width=\linewidth]{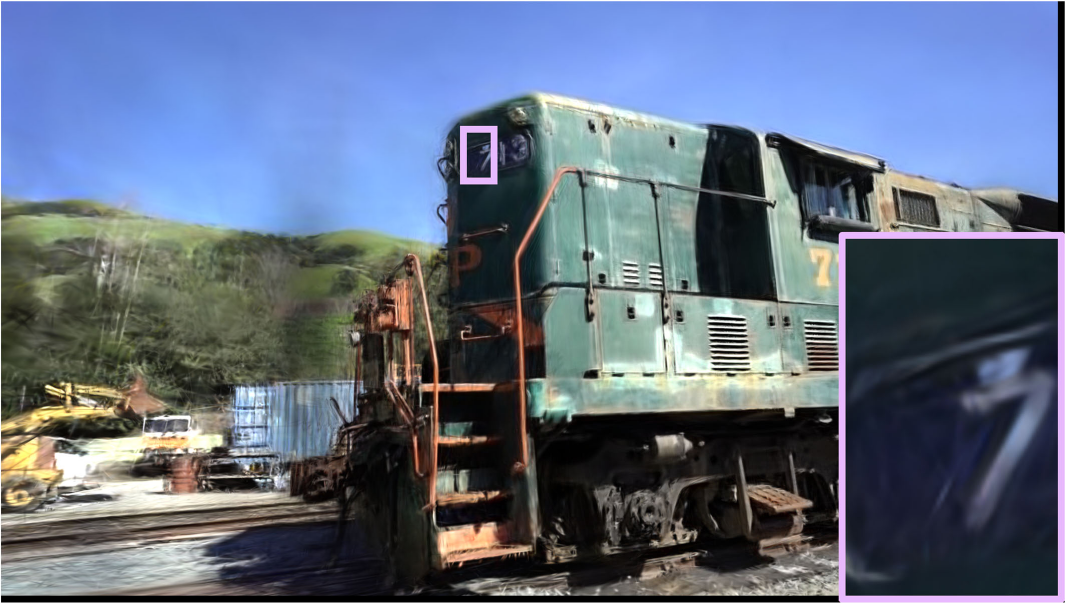}
\end{minipage}\hfill
\begin{minipage}[b]{0.33\textwidth}\centering
\includegraphics[width=\linewidth]{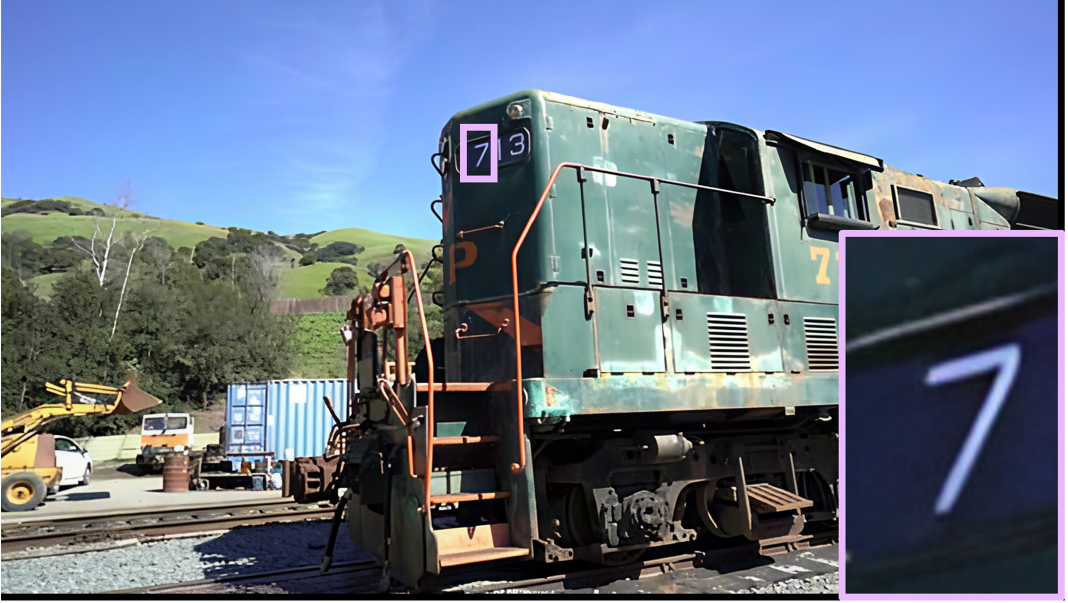}
\end{minipage}

\vspace{1mm}

\begin{minipage}[b]{0.33\textwidth}\centering {w/o $X_t$}\end{minipage}\hfill
\begin{minipage}[b]{0.33\textwidth}\centering {FULL}\end{minipage}\hfill
\begin{minipage}[b]{0.33\textwidth}\centering {GT}\end{minipage}

\vspace{1mm}

\begin{minipage}[b]{0.33\textwidth}\centering
\includegraphics[width=\linewidth]{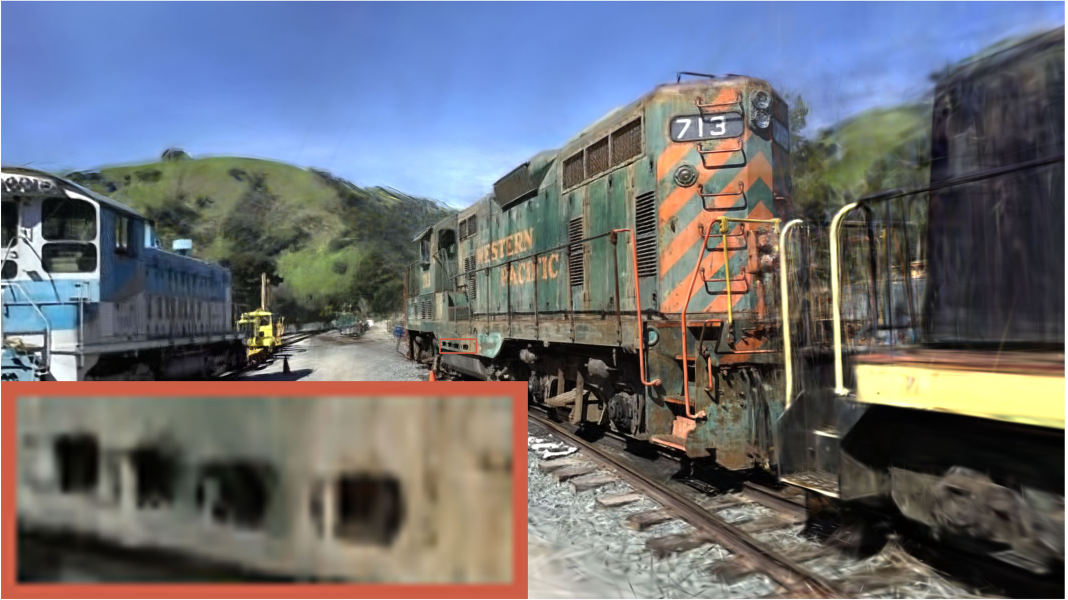}
\end{minipage}\hfill
\begin{minipage}[b]{0.33\textwidth}\centering
\includegraphics[width=\linewidth]{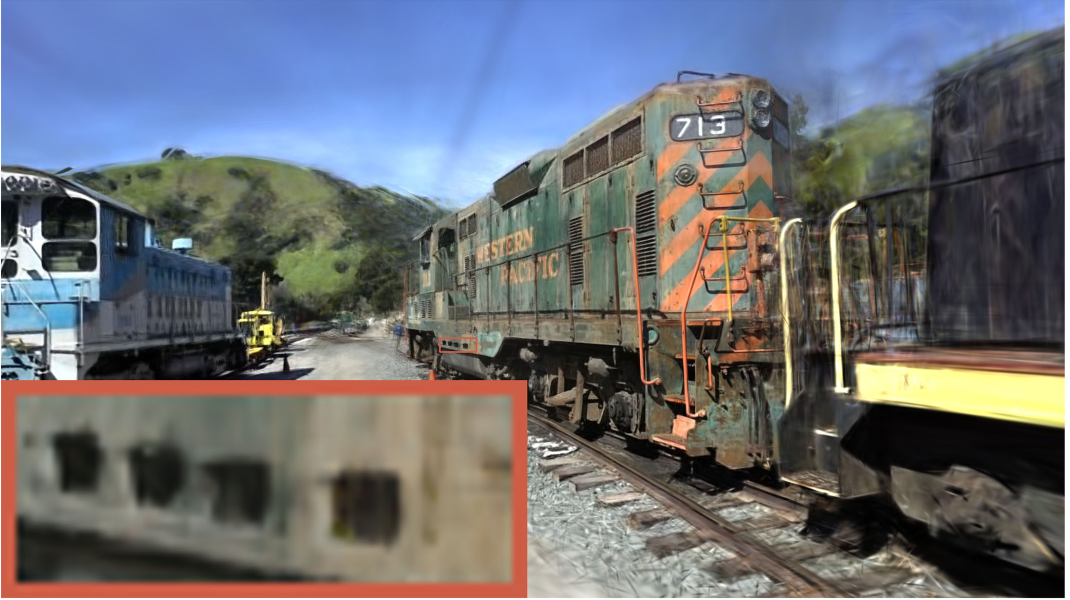}
\end{minipage}\hfill
\begin{minipage}[b]{0.33\textwidth}\centering
\includegraphics[width=\linewidth]{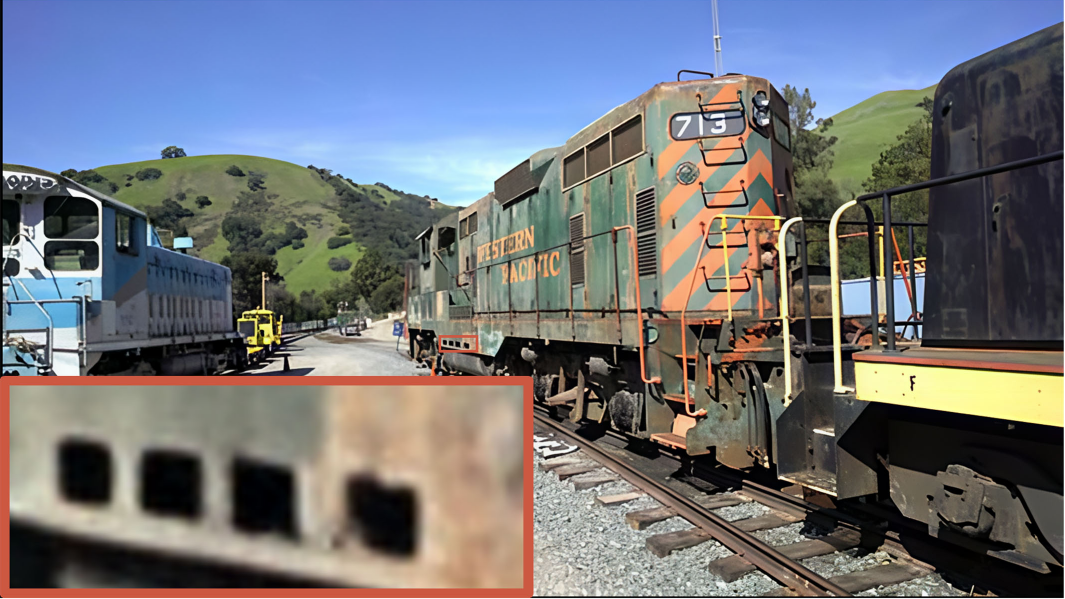}
\end{minipage}

\caption{Qualitative analysis of the component-wise removal ablation on the frequency-aware reliability map $C^{rel}_{t}$ using the train scene from the Tanks \& Temples dataset. The three comparison groups show the effects of removing $E_t$, $G_t$, and $X_t$, respectively. Removing $E_t$ weakens structural alignment, removing $G_t$ weakens the recovery of unresolved high-frequency details, and removing $X_t$ allows more view-inconsistent pseudo details to remain in the reconstruction.}
\label{fig:ablation_reliability_terms}
\end{figure*}

Table~\ref{tab:ablation_reliability_components} shows that removing any cue degrades performance. Without $E_t$, details are injected into structurally unsupported regions; without $G_t$, unresolved spectral discrepancy is ignored; without $X_t$, view-inconsistent pseudo-details harm multi-view stability. The full model achieves the best results (PSNR 23.42, SSIM 0.775, LPIPS 0.281), confirming that reliable detail injection requires all three cues.

\begin{table}[t]
\centering
\caption{Component-wise ablation on the reliability map $C^{rel}_{t}$ (train scene, Tanks \& Temples).}
\label{tab:ablation_reliability_components}
\setlength{\tabcolsep}{6pt}
\renewcommand{\arraystretch}{1.12}
\begin{tabular*}{\columnwidth}{@{\hspace{0.8em}\extracolsep{\fill}}lccc@{\hspace{0.8em}}}
\hline
Variant & PSNR$\uparrow$ & SSIM$\uparrow$ & LPIPS$\downarrow$ \\
\hline
w/o $E_t$ & 23.32 & 0.758 & 0.294 \\
w/o $G_t$ & 23.09 & 0.757 & 0.294 \\
w/o $X_t$ & 23.18 & 0.758 & 0.293 \\
Full      & \textbf{23.42} & \textbf{0.775} & \textbf{0.281} \\
\hline
\end{tabular*}
\end{table}

We also compare the design of $G_t$. A spatial-gradient variant replaces the Fourier high-pass residual with a Sobel-gradient residual:
\begin{equation}
G_t^{grad} = \operatorname{Norm}\left(\left||\nabla R_t^{HR}| - |\nabla I_t^{SR}|\right|\right)
\end{equation}
The full model uses the Fourier-domain high-pass residual:
\begin{equation}
G_t = \operatorname{Norm}\left(\left|\mathcal{H}(R_t^{HR}) - \mathcal{H}(I_t^{SR})\right|\right)
\end{equation}
where $\mathcal{H}(\cdot)$ denotes the Fourier-domain high-pass operator.

Table~\ref{tab:ablation_gt_design} shows that the spatial-gradient variant improves over removing $G_t$ (PSNR 23.24 vs. 23.09), but the Fourier high-pass residual achieves the best performance (PSNR 23.42). Spatial gradients are more sensitive to noise and misalignment, while the Fourier-domain residual directly measures unresolved spectral discrepancy, providing a cleaner cue for reliability estimation.

\begin{table}[t]
\centering
\caption{Ablation on the $G_t$ design (train scene, Tanks \& Temples).}
\label{tab:ablation_gt_design}
\setlength{\tabcolsep}{6pt}
\renewcommand{\arraystretch}{1.12}
\begin{tabular*}{\columnwidth}{@{\hspace{0.8em}\extracolsep{\fill}}lccc@{\hspace{0.8em}}}
\hline
Variant & PSNR$\uparrow$ & SSIM$\uparrow$ & LPIPS$\downarrow$ \\
\hline
w/o $G_t$ & 23.09 & 0.757 & 0.294 \\
Spatial-gradient $G_t$ & 23.24 & 0.767 & 0.286 \\
Fourier high-pass $G_t$ (Full) & \textbf{23.42} & \textbf{0.775} & \textbf{0.281} \\
\hline
\end{tabular*}
\end{table}

\subsection{Ablation on Reliability-Aware Gaussian Densification}

We evaluate the densification score $S_i = (\bar{U}_i \bar{F}_i \bar{H}_i)^{1/3}$ on the train scene from Tanks \& Temples. Table~\ref{tab:ablation_densification_terms} shows that removing any term degrades performance. Removing $U_i$ causes a modest drop (PSNR 23.32), removing $F_i$ causes a larger drop (PSNR 23.10), and removing $H_i$ causes the largest degradation (PSNR 22.57, LPIPS 0.438). This confirms that effective densification requires joint consideration of geometric demand, reliable spectral residuals, and local structural complexity.

The relative impact of each densification term reveals how different cues guide Gaussian proliferation. The geometric demand term $U_i$ ensures that densification is concentrated in regions where the current representation is genuinely insufficient, preventing wasteful cloning in already well-reconstructed areas. The spectral residual term $F_i$ directs new Gaussians toward regions where reliable high-frequency content remains unresolved, aligning capacity expansion with the most informative supervision signals. The structural complexity term $H_i$ prevents over-smoothing in textured regions by promoting densification where local patterns require finer geometric granularity. The particularly strong degradation when $H_i$ is removed (LPIPS rises to 0.438) suggests that local structural complexity is essential for preserving perceptual quality, as naive densification without texture awareness tends to produce uniform, overly smooth reconstructions that lose fine surface details.

\begin{table}[t]
\centering
\caption{Ablation on the densification score components (train scene, Tanks \& Temples).}
\label{tab:ablation_densification_terms}
\setlength{\tabcolsep}{6pt}
\renewcommand{\arraystretch}{1.12}
\begin{tabular*}{\columnwidth}{@{\hspace{1.0em}\extracolsep{\fill}}cccc@{\hspace{1.0em}}}
\hline
Variant & PSNR$\uparrow$ & SSIM$\uparrow$ & LPIPS$\downarrow$ \\
\hline
w/o $U_i$ & 23.32 & 0.758 & 0.294 \\
w/o $F_i$ & 23.10 & 0.755 & 0.296 \\
w/o $H_i$ & 22.57 & 0.725 & 0.438 \\
Full      & \textbf{23.42} & \textbf{0.775} & \textbf{0.281} \\
\hline
\end{tabular*}
\end{table}

\subsection{Effect of the Number of Selected Frequency Patches}

We study the number of selected frequency patches $K$ on the train scene (Tanks \& Temples) and bicycle scene (Mip-NeRF 360), averaging five runs per setting. Table~\ref{tab:patch_number} shows that $K=2$ provides a good trade-off: it improves over $K=0$ (PSNR 23.31 vs. 23.23) with minimal time increase (0.74 h vs. 0.72 h). Larger $K$ brings diminishing gains while significantly increasing training time ($K=16$ takes 1.78 h). Therefore, $K=2$ is used as the default.

\begin{figure}[!t]
\centering
\includegraphics[width=3in]{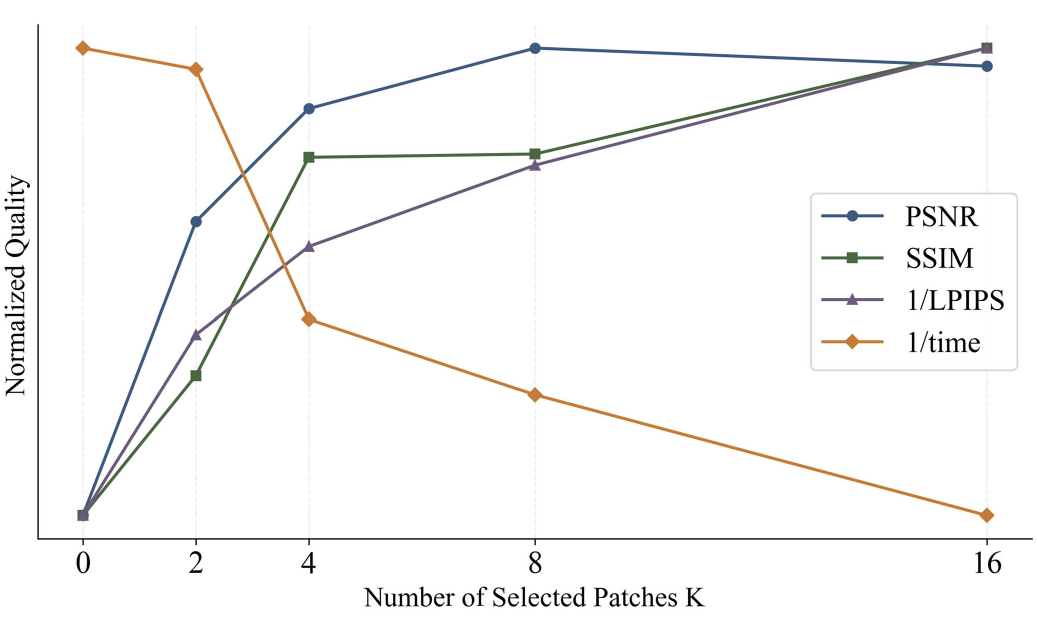}
\caption{Normalized quality-efficiency trade-off under different $K$ values (averaged over train and bicycle scenes). Higher values indicate better performance.}
\label{fig:patch_number}
\end{figure}

The diminishing returns with larger $K$ can be attributed to the spatial redundancy among high-frequency patches. When a scene contains large uniform regions or regions with consistent texture patterns, additional patches often sample similar spectral content and provide redundant gradient information rather than complementary constraints. Moreover, each additional patch requires a separate forward pass through the Fourier analysis pipeline and increases the memory footprint during training, making the cost grow super-linearly with $K$. The efficiency analysis shown in Figure~\ref{fig:patch_number} further supports this observation: the normalized quality score saturates quickly after $K=2$, while the normalized time cost continues to increase steadily. This indicates that a small number of carefully selected patches is sufficient to inject reliable frequency guidance without incurring prohibitive computational overhead.

\begin{table}[t]
\centering
\caption{Effect of the number of selected frequency patches $K$ (averaged over train and bicycle scenes).}
\label{tab:patch_number}
\setlength{\tabcolsep}{6pt}
\renewcommand{\arraystretch}{1.12}
\begin{tabular*}{\columnwidth}{@{\hspace{1.5em}\extracolsep{\fill}}ccccc@{\hspace{1.5em}}}
\hline
$K$ & PSNR$\uparrow$ & SSIM$\uparrow$ & LPIPS$\downarrow$ & Time$\downarrow$ \\
\hline
0  & 23.23 & 0.773 & 0.282 & \textbf{0.72} h \\
2  & 23.31 & 0.776 & 0.279 & 0.74 h \\
4  & 23.35 & 0.781 & 0.277 & 1.10 h \\
8  & 23.35 & 0.781 & 0.276 & 1.29 h \\
16 & \textbf{23.36} & \textbf{0.783} & \textbf{0.274} & 1.78 h \\
\hline
\end{tabular*}
\end{table}

\subsection{Discussion}

The ablation studies confirm that the proposed components play complementary roles: demand estimation locates under-detailed regions, reliability assessment filters unreliable candidates, progressive regularization stabilizes optimization, and densification allocates capacity to unresolved reliable structures. The full design consistently outperforms reduced alternatives across all tested configurations.

Furthermore, the systematic degradation observed when removing individual components suggests that none of the proposed mechanisms is redundant. Each addresses a distinct failure mode of naive SR-guided 3DGS: demand estimation prevents wasteful supervision in well-sampled regions, reliability assessment blocks view-inconsistent hallucinations, progressive scheduling avoids premature high-frequency enforcement, and targeted densification ensures sufficient representation capacity. These complementary safeguards collectively shift the reconstruction objective from maximizing local image sharpness toward building a structurally stable and multi-view consistent 3D representation.

\section{CONCLUSION}
\label{sec:CONCLUSION}

We presented a reliability-aware frequency modeling framework for low-resolution multi-view 3D Gaussian Splatting. Instead of directly amplifying fine details from low-resolution inputs or external super-resolution references, our method selectively integrates candidate frequency cues only when they are both needed and supported by reliable evidence. The proposed framework combines geometry-guided detail-demand estimation, frequency-aware reliability assessment, selective progressive frequency regularization, and reliability-aware Gaussian densification. Geometry cues locate under-resolved regions, while structural, spectral, and cross-view consistency signals evaluate whether candidate high-frequency content reflects real scene details rather than hallucinated or view-dependent artifacts. This reliability guidance modulates both frequency-domain supervision and Gaussian densification, allowing representation capacity to be allocated progressively to trustworthy unresolved structures.

Experiments on multiple benchmark datasets demonstrate improved reconstruction fidelity and perceptual quality under low-resolution settings. Ablation studies further confirm the complementary roles of the proposed components in locating missing details, filtering unreliable cues, stabilizing supervision, and guiding densification. Limitations remain, including dependence on external super-resolution quality and additional training-time computation. Future work may explore lightweight reliability prediction and more efficient update or densification strategies. Overall, this work emphasizes evidence-aware detail integration for robust low-resolution 3DGS.

\section*{Acknowledgments}
This should be a simple paragraph before the References to thank those individuals and institutions who have supported your work on this article.

{


\newpage

\end{document}